\documentclass[5p]{elsarticle}

\usepackage{amsmath,amsfonts}
\usepackage{algorithmic}
\usepackage{array}
\usepackage[caption=false,font=normalsize,labelfont=sf,textfont=sf]{subfig}
\usepackage{textcomp}
\usepackage{stfloats}
\usepackage{url}
\usepackage{verbatim}
\usepackage{graphicx}
\usepackage{enumitem}
\usepackage{xcolor}

\usepackage[T1]{fontenc}
\usepackage{amssymb}
\usepackage{booktabs}

\usepackage{pifont}

\usepackage[utf8]{inputenc}
\usepackage{multicol, multirow}
\usepackage{booktabs}
\usepackage{bm}

\usepackage{flushend}

\usepackage{color, colortbl}

\usepackage{contour}
\contourlength{1pt}

\usepackage{overpic}

\usepackage{pgf,tikz}
\usepackage{tkz-euclide,tkz-graph}
\usetikzlibrary{calc}
\usepackage{pgfplots}
\pgfplotsset{
    compat=1.3,
    tick label style={font=\footnotesize},
    label style={font=\footnotesize},
    width=9cm,
    height=6cm}

\newcommand{\cmark}{\ding{51}}

\definecolor{mygray}{gray}{0.9} 
\newcolumntype{g}{>{\columncolor{mygray}}c}

\definecolor{box}{RGB}{72, 118, 177}
\definecolor{can}{RGB}{228, 126, 35}
\definecolor{banana}{RGB}{97, 159, 58}
\definecolor{powerdrill}{RGB}{184, 44, 44}
\definecolor{scissors}{RGB}{137, 103, 185}
\definecolor{pear}{RGB}{127, 87, 77}
\definecolor{dish}{RGB}{202, 119, 190}
\definecolor{camel}{RGB}{126, 126, 126}
\definecolor{mouse}{RGB}{187, 188, 57}
\definecolor{shampoo}{RGB}{109, 189, 206}

\definecolor{head}{RGB}{178, 205, 225}
\definecolor{hands}{RGB}{73, 119, 177}
\definecolor{forearms}{RGB}{191, 222, 142}
\definecolor{upperarms}{RGB}{99, 159, 58}
\definecolor{torso}{RGB}{228, 153, 152}
\definecolor{thighs}{RGB}{195, 32, 34}
\definecolor{legs}{RGB}{237, 189, 115}
\definecolor{foot}{RGB}{228, 126, 29}
\definecolor{correct}{RGB}{244, 232, 242}
\definecolor{wrong}{RGB}{219, 33, 33}

\definecolor{pn}{RGB}{106, 61, 154}
\definecolor{pn2}{RGB}{31, 120, 180}
\definecolor{dgcnn}{RGB}{51, 160, 44}
\definecolor{pt}{RGB}{245, 216, 69}
\definecolor{pointmixer}{RGB}{255, 127, 0}
\definecolor{ours}{RGB}{227, 26, 28}

\begin{document}\sloppy

\begin{frontmatter}

\title{PatchMixer: Rethinking network design to boost generalization\\ for 3D point cloud understanding}

\author[1]{Davide Boscaini\corref{cor1}}
\ead{dboscaini@fbk.eu}

\author[1]{Fabio Poiesi}
\ead{poiesi@fbk.eu}

\cortext[cor1]{Corresponding author}

\affiliation[1]{
    organization={Technologies of Vision Lab, Fondazione Bruno Kessler},
    addressline={via Sommarive 18},
    postcode={38123},
    city={Trento},
    country={Italy}}

\begin{abstract}
The recent trend in deep learning methods for 3D point cloud understanding is to propose increasingly sophisticated architectures either to better capture 3D geometries or by introducing possibly undesired inductive biases.
Moreover, prior works introducing novel architectures compared their performance on the same domain, devoting less attention to their generalization to other domains.
We argue that the ability of a model to transfer the learnt knowledge to different domains is an important feature that should be evaluated to exhaustively assess the quality of a deep network architecture.
In this work we propose PatchMixer, a simple yet effective architecture that extends the ideas behind the recent MLP-Mixer paper to 3D point clouds.
The novelties of our approach are the processing of local patches instead of the whole shape to promote robustness to partial point clouds, and the aggregation of patch-wise features using an MLP as a simpler alternative to the graph convolutions or the attention mechanisms that are used in prior works.
We evaluated our method on the shape classification and part segmentation tasks, achieving superior generalization performance compared to a selection of the most relevant deep architectures.
\end{abstract}


\end{frontmatter}

\section{Introduction}\label{sec:intro}

Point cloud understanding is important for real-world applications including
brain connectivity analysis~\cite{TractogramFiltering},
protein interactions prediction~\cite{ProteinInteractions},
biometric recognition~\cite{ClusteredDGCNN},
robotic manipulation~\cite{GeDi},
and autonomous driving~\cite{loopZhou2022}.
Unlike 2D images, that are regular grids of pixels, point clouds are sparse, unordered, and irregular sets of 3D points~\cite{PN}.
This structural difference hinders the application of conventional convolutional neural networks developed for 2D image processing to 3D point clouds.

Early approaches addressed this issue by introducing different preprocessing strategies to convert 3D surfaces to regular representations. 
Multi-View CNN~\cite{Su2015} represents a 3D shape as a collection of 2D rendered views and processes them with a CNN using 2D convolutional layers.
3D ShapeNets~\cite{Wu2015} represents a 3D shape as a volumetric occupancy grid of binary voxels and processes it with a Convolutional Deep Belief Network using 3D convolutional layers,
Qi et al.~\cite{Qi2016} proposes a novel volumetric CNN design that performs on par with multi-view approaches and identifies the 3D resolution as the main bottleneck for the performance of volumetric CNNs,
OctreeCNN~\cite{Wang2017} addresses such bottleneck by exploiting octree representations.

Lately, a new trend emerged where the focus is shifted towards the design of end-to-end deep architectures processing point clouds directly, without requiring preprocessing.
These approaches require the development of new building blocks replacing traditional deep layers.
Point-based approaches employ permutation-invariant operators such as point-level MLPs and pooling layers to aggregate features from single or multiple scales~\cite{PN, PN2}.
Convolution-based approaches perform convolution on 3D deformable surfaces as template matching between learnable filters and local patches~\cite{LSCNN, GCNN, ADD, ACNN, MoNet, pointcnn, shellnet, riconv}.
Graph-based approaches perform message passing on local graphs connecting neighboring points~\cite{ECC, GAT, DGCNN}.
Attention-based approaches merge local features using self-attention and positional encoding~\cite{PointTransformer, Engel2021, Guo2021}.

These approaches are designed to improve the recognition of fine geometric patters.
As a result, the mainstream in designing novel architectures for point cloud analysis is to develop increasingly sophisticated local feature extractors~\cite{PointMLP}, which are typically trained and tested on disjoint splits of the same data distribution.
Popular benchmarks include ModelNet40~\cite{Wu2015} for classification and ShapeNet-Part~\cite{ShapeNetPart} for part segmentation.
We argue that performance evaluation across same domains does not suffice to comprehensively assess the effectiveness of a model because:
i) such datasets have limited cardinality and their training and test distributions are similar, and
ii) performances are nearly saturated~\cite{PointMLP}.
On the contrary, one can perform comparative experiments by assessing the generalization ability of different architectures to different domains.

\begin{figure}[t]

    \hspace{2mm}
    \begin{overpic}[width=0.97\columnwidth, trim=0 0 0 0, clip]{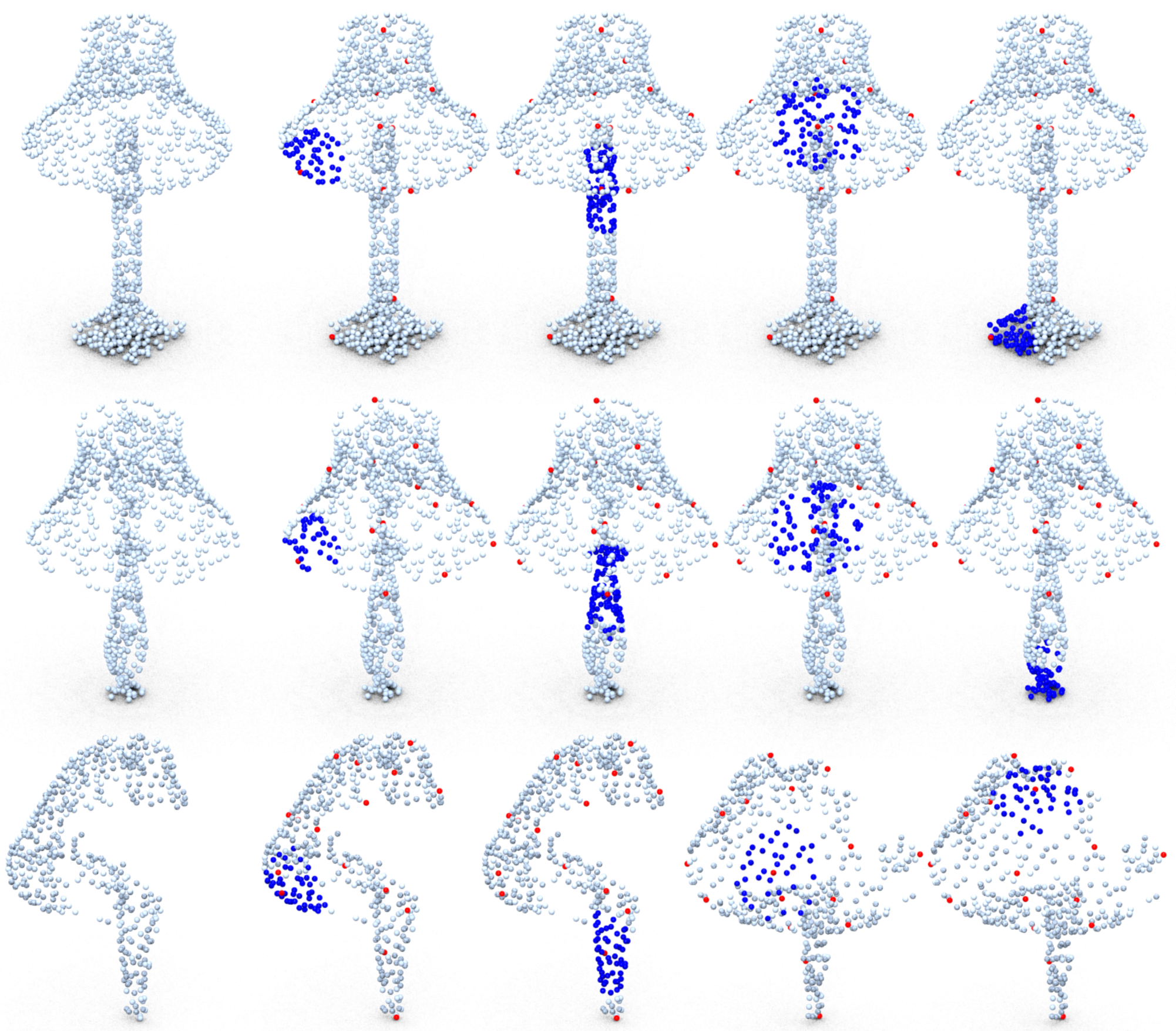}
        \put(-3,6){\footnotesize \rotatebox{90}{ScanNet}}
        \put(-3,33){\footnotesize \rotatebox{90}{ShapeNet}}
        \put(-3,63){\footnotesize \rotatebox{90}{ModelNet}}
        \put(21.5,0){\color{black}\rule{0.33pt}{214pt}}
    \end{overpic}

    \vspace{-4mm}
    \caption{
    Point clouds of lamps from different subsets of PointDA dataset~\cite{PointDAN}.
    Left, light blue: Shapes show intra-class variations (different bases) and  acquisition artifacts (partiality).
    Right, dark blue: Patches extracted from such shapes highlight similar (columns 1--4) or dissimilar (column 5) structures. 
    }
    \label{fig:patches}
\end{figure}
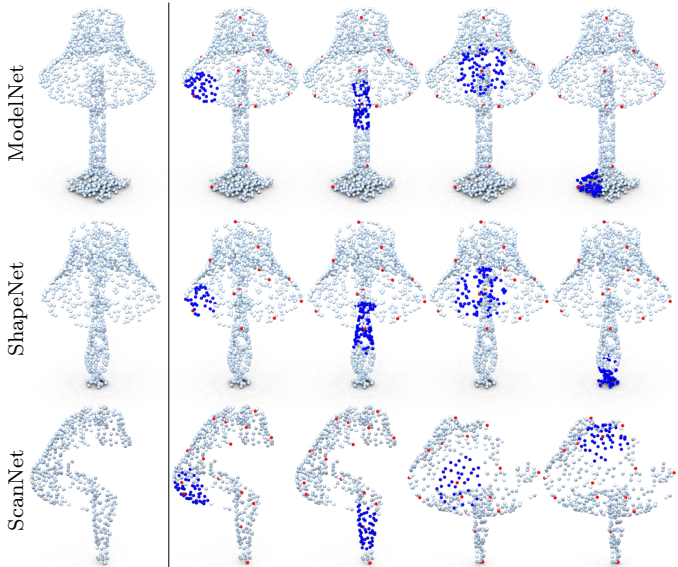

In this paper, we propose a simple yet effective architecture that extends the MLP-Mixer architecture~\cite{MLPMixer, ConvMixer} from the analysis of 2D images to 3D point clouds.
Our method relies on raw 3D data (point coordinates), aims to remove inductive biases, and is designed by following two main criteria: locality and simplicity.
\textit{Locality}:
We decompose the input shape as a collection of local patches, extract patch-level features, and learn global representations by mixing information across patches.
This allows us to achieve robustness to intra-class variations and acquisition noise such as missing parts due to occlusions.
In Fig.~\ref{fig:patches} we show an example where each row contains a lamp instance from different subsets of the PointDA dataset~\cite{PointDAN}.
Globally, the shapes on the left-hand side are different, e.g.~the top shape has a different base than the other two, the bottom shape is instead incomplete.
However, locally (at patch level) we can easily identify similarities.
\textit{Simplicity}:
Instead of designing a novel local feature extractor more sophisticated than competitors, we employ only simple and highly efficient layers such as MLPs, batch normalisations, ReLUs, and residual connections.
The main novelty introduced by our architecture is the learnable patch-level feature aggregation performed by an attentive token mixer module.
This aggregation function is an MLP and is optimized during training and contextually to the feature extractor to solve a downstream task.
We name our method \emph{PatchMixer} and its implementation is publicly available at \url{https://github.com/davideboscaini/PatchMixer}.

\begin{figure}[t!]
    \begin{tikzpicture}
        \begin{axis}[
            title={\small Real RealSense $\to$ Real Kinect},
            xlabel={\small Number of parameters [M]},
            ylabel={\small Overall Accuracy [\%]},
            ymajorgrids=true,
            xmajorgrids=true,
            scatter,
            only marks,
            mark size=4pt
        ]
        \addplot[mark=square*, mark options={scale=0.9}] table{
            x y
            0.809802 79.2
        };
        \node[right] at (4, -1) {\small \contour{white}{PointNet \cite{PN}}};
        \addplot[mark=halfsquare*, mark options={rotate=-45, scale=1.25}] table{
            x y
            1.467210 82.2
        };
        \node[right] at (-14, 42) {\small \contour{white}{PointNet++ \cite{PN2}}};
        \addplot[mark=triangle*, mark options={scale=1.25}] table{
            x y
            1.801610 82.0
        };
        \node[right] at (14, 24) {\small \contour{white}{DGCNN \cite{DGCNN}}};
        \addplot[mark=pentagon*, mark options={scale=1}] table{
            x y
            9.684536 89.8
        };
        \node[right] at (55, 96) {\small \contour{white}{PointTransformer \cite{PointTransformer}}};
        \addplot[mark=diamond*, mark options={scale=1.1}] table{
            x y
            13.230000 91.4
        };
        \node[right] at (87, 132) {\small \contour{white}{PointMLP \cite{PointMLP}}};
        \addplot[mark=halfdiamond*, mark options={scale=1.1}] table{
            x y
            4.140000 84.7
        };
        \node[right] at (20, 65) {\small \contour{white}{PointMixer \cite{PointMixer}}};
        \addplot[mark=*, mark options={scale=1.25}] table{
            x y
            2.319370 93.6
        };
        \node[right] at (19, 145) {\small \contour{white}{PatchMixer (Ours)}};
        \end{axis} 
    \end{tikzpicture}

    \vspace{-5mm}
    \caption{
        Comparison of shape classification performance of different models.
        The graph relates the number of learnable parameters with the Overall Accuracy reached in a transfer learning scenario (when training on Real RealSense and testing on Real Kinect splits of the GraspNetPC dataset~\cite{GraspNet}).
        A common trend in the competitors design emerges: increasing the network capacity leads to a performance improvement.
        Our approach breaks away from this trend: PathMixer is the most accurate method while requiring a significantly lower number of parameters.
    }
    \label{fig:teaser}
\end{figure}
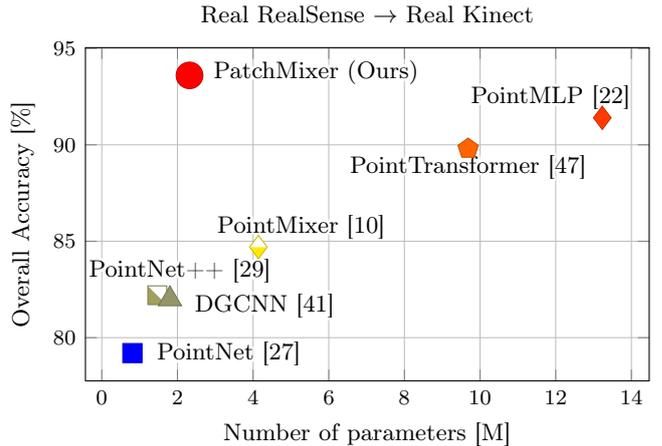

We extensively evaluate PatchMixer on classification and part segmentation tasks.
Classification is evaluated on the GraspNetPC~\cite{GAI} and PointDA~\cite{PointDAN} benchmarks.
Part segmentation is evaluated on the PointSegDA benchmark~\cite{DefRec}.
We compare our method against a selection of the most relevant deep architectures processing point clouds directly, i.e. PointNet~\cite{PN}, PointNet++~\cite{PN2}, Dynamic Graph CNN~\cite{DGCNN}, PointTransformer~\cite{PointTransformer}, PointMLP~\cite{PointMLP}, and PointMixer~\cite{PointMixer}.
PatchMixer outperforms these methods in transfer learning scenarios, by a large margin in some configurations, while also achieving comparable performance in the same domain scenarios.

To summarize, our contributions are:
\setlist{nolistsep}
\begin{itemize}
    \item a novel deep learning architecture to process 3D point cloud data based on MLP layers only in order to minimize inductive biases;
    \item a learnable patch-level feature aggregation mechanism that is performed by an attentive token mixer module;
    \item a comprehensive transfer learning evaluation that, to the best of our knowledge, is missing in the related literature.
\end{itemize}

The paper is organized as follows. 
Sec.~\ref{sec:related} describes the related work. 
Sec.~\ref{sec:approach} presents the formulation of our approach.
Sec.~\ref{sec:experiments} includes the experiments and the comparisons to validate our approach. 
Sec.~\ref{sec:conclusions} draws the conclusions.

\section{Related work}\label{sec:related}
Deep learning backbones for 3D point cloud data can be grouped into five categories: 
point-based,
convolution-based,
graph-based,
attention-based,
and MLP-based.
A survey of other deep learning backbones is also available~\cite{survey}.

\subsection{Point-based methods}
Point-based methods takes as input the extrinsic coordinates of the points and outputs local features by processing each point independently.
PointNet (PN)~\cite{PN} is a end-to-end architecture that employs only point-wise MLPs and symmetric functions to achieve invariance with respect to input point permutations.
Shape classification is performed by aggregating local features through max pooling.
Part segmentation is performed by concatenating the aggregated local features (global feature) to the local features.
Despite being light, fast and versatile, PN has two limitations:
i) because it operates on each point independently, it cannot effectively capture the local geometric context around each point, and
ii) because it operate at a single scale, it cannot learn hierarchical abstractions for the generalization to unseen cases.
PointNet++ (PN2)~\cite{PN2} extends PN by implementing a grouping mechanism to extract features at different scales.
This enables PN2 to improve robustness to non-uniform point densities.
At each scale, PN2 uses PN as feature extractor to achieve invariance to point permutations.
Unlike PN, the features learned with PN2 capture different levels of detail, enabling generalization to unseen data.
Two variants of PN2 can be used:
i) Single Scale Grouping (SSG), which operates at a single scale, and
ii) Multi Scale Grouping (MSG), which operated at multiple scales.
PN2 shares a limitation with PN, i.e.~at a given scale it still processes points independently, failing to capture the geometric relations among neighborhoods.

\subsection{Convolution-based methods}
Convolution-based methods extend the conventional convolution operation to non-Euclidean manifolds.
They can be divided into spectral and spatial approaches depending on the domain where the convolution is performed.
Spectral methods exploit the Convolution Theorem to compute the convolution of two signals as a point-wise product of their Fourier transforms~\cite{Bruna2014, LSCNN, Defferrard2016, Kipf2016}.
Spectral CNNs~\cite{Bruna2014} defines convolutional layers on graphs with a fixed topology by computing the Fourier transform through the eigendecomposition of the graph Laplacian.
LSCNN~\cite{LSCNN} introduces a generalization of the windowed Fourier transform to represent the local context around a point on a surface in the frequency domain and learns filters over such representation with a MLP.
ChebNet~\cite{Defferrard2016} improves the efficiency of spectral CNNs by avoiding the explicit computation of the Laplacian eigenvectors in favor of recurrent Chebyshev polynomials.
GCN~\cite{Kipf2016} further simplified ChebNet using simpler filters operating on more local neighborhoods.
Spatial methods learns context-aware filters over local 3D patches similarly to how traditional CNNs learns convolutional filters over local 2D grids of pixels~\cite{GCNN, ADD, ACNN, MoNet}.
These approaches require the definition of a patch operator mapping a local patch around a point to a grid-like representation via local charting.
Different definitions of the patch operator leads to different methods.
GCNN~\cite{GCNN} introduces a patch operator crafted for triangular meshes representing local 3D patches in a geodesic polar coordinate system and learns 2D filters over such representation with a MLP.
ACNN~\cite{ADD, ACNN} computes the local charting by projecting the signal defined on the 3D surface into a collection of anisotropic heat kernels acting as soft bins of a local polar coordinate system.
MoNet~\cite{MoNet} proposes to learn the weighting functions of the patch operator as Gaussian Mixtures over a local pseudo-coordinate system, resulting in a generic framework that includes GCNN and ACNN as particular cases using hand-crafted weighting functions.
PointCNN~\cite{pointcnn} further relaxes the local charting assumption of patch operator-based approaches~\cite{GCNN, ADD, ACNN, MoNet}, by mapping the input points to a latent space by means of an $\mathcal{X}$-transformation that learns how to weight the input features and how to permute the input points to achieve a canonical ordering.
Convolution over such latent space is performed using element-wise product and sum operations.
Rather than mapping the input points to a  latent space, ShellNet~\cite{shellnet} performs convolution-like operations directly in the input space by applying MLPs over local representations obtained from multi-scale concentric spherical shell statistics.
Differently than competitor designs, ShellNet can increase its receptive field even without stacking layers, thus saving learnable parameters.
ShellNet achieves point permutation invariance through max-pooling operations, as in PN~\cite{PN}.
Spatial resolution is decreased across layers by means of point sampling, as in PN2~\cite{PN2}.
RIConv~\cite{rotinvconv} and its improved version, RIConv++~\cite{riconv}, introduce a convolution operator that is also invariant to input shape rotations.
This is achieved by extracting low-level geometric features such as input point distances and angles to obtain a local representation similar to the local charting created by patch operator-based approaches~\cite{GCNN, ADD, ACNN, MoNet}.
Such representation is first lifted to an higher-dimensional feature space by means of MLPs, and then projected along a local reference axis to obtain an 1D histogram representation, over which 1D convolution is performed.
Differently than PN~\cite{PN}, invariance to the input point ordering is enforced by the binning procedure creating the 1D histogram representation.
Similarly to PN2~\cite{PN2}, downsampling and upsampling are used to manipulate the spatial resolution of the input shape.
Rotation invariance comes to the expense of a distinctiveness reduction~\cite{rotinvconv, riconv}.

\subsection{Graph-based methods}
Graph-based methods apply graph neural networks to graph representations of point clouds, where nodes are represented by the vertices and edges connect neighboring points~\cite{ECC, GAT, DGCNN}.
By applying convolution-like operations on local neighborhood graphs, these approaches can encode local geometric structures instead of working on individual points like PN~\cite{PN} or PN2~\cite{PN2}.
Dynamic Graph CNN (DGCNN)~\cite{DGCNN} is the most prominent example of this category.
The main difference between DGCNN and traditional graph CNNs~\cite{ECC, GAT} is the dynamic update performed at each layer of the network as opposed to having a fixed proximity graph: 
at the first layer the graph connects spatially neighboring points, while at deeper layers it connects nodes based on their feature similarity.
This dynamic computation of the proximity graph allows DGCNN to aggregate features from potentially distant vertices across the point cloud.
Both PN and PN2 can be thought as special cases of DGCNN:
PN can be reproduced when considering a proximity graph with an empty edge set;
PN2 can be reproduced by first building a fixed proximity graph according to the Euclidean distance between points, and then applying a graph coarsening operation to create the different levels of the hierarchy.
Unlike PN and PN2, DGCNN can incorporate local neighborhood information while maintaining permutation invariance.

\subsection{Attention-based methods}
Attention-based methods include PCT~\cite{Guo2021} and PointTransformer (PT)~\cite{PointTransformer, Engel2021}.
PT extends the conventional Transformer architecture~\cite{Vaswani2017} to 3D point cloud processing.
PT introduces a self-attention mechanism to aggregate the local features extracted with point-level operations.
The self-attention operator is inherently suited to process point clouds because it is invariant to permutation and cardinality of the input set.
PT uses a learnable positional encodings to integrate point positional information in the model.
One limitation of PT is that its self-attention layers require quadratic runtime.

\subsection{MLP-based methods}
MLP-based methods include PointMLP~\cite{PointMLP} and PointMixer~\cite{PointMixer}.
These methods are based on the MLP-Mixer~\cite{MLPMixer} and ConvMixer~\cite{ConvMixer} architectures that aims to remove inductive biases from models and rely on raw data.
PointMLP introduces a geometric affine module to normalize the input data statistics. 
PointMixer removes the token mixer layer and integrates a layer to aggregate features across different sets (intra-set, inter-set and hierarchical-set) of points.
The two main differences between PatchMixer and these two approaches are the network design and the objective.
PatchMixer modifies the original MLP-Mixer~\cite{MLPMixer} design by introducing the attentive token mixer.
Our model design instead avoids the introduction of any inductive bias on purpose and uses only conventional and highly-optimized layers such as MLPs, skip connections, normalization layers, and nonlinear activations.
Both PointMLP and PointMixer are evaluated on the same domain scenario only, achieving marginal gains compared to prior works.
Instead, we compare PatchMixer to the other networks designs in terms of transfer learning.
To the best of our knowledge, this is the first work that comprehensively analyzes how MLP-based architectures generalize across 3D point cloud distributions.

\section{Our approach}\label{sec:approach}

\captionsetup{skip=0pt}
\begin{figure*}[t]
    \centering
    \begin{overpic}[width=1.0\textwidth]{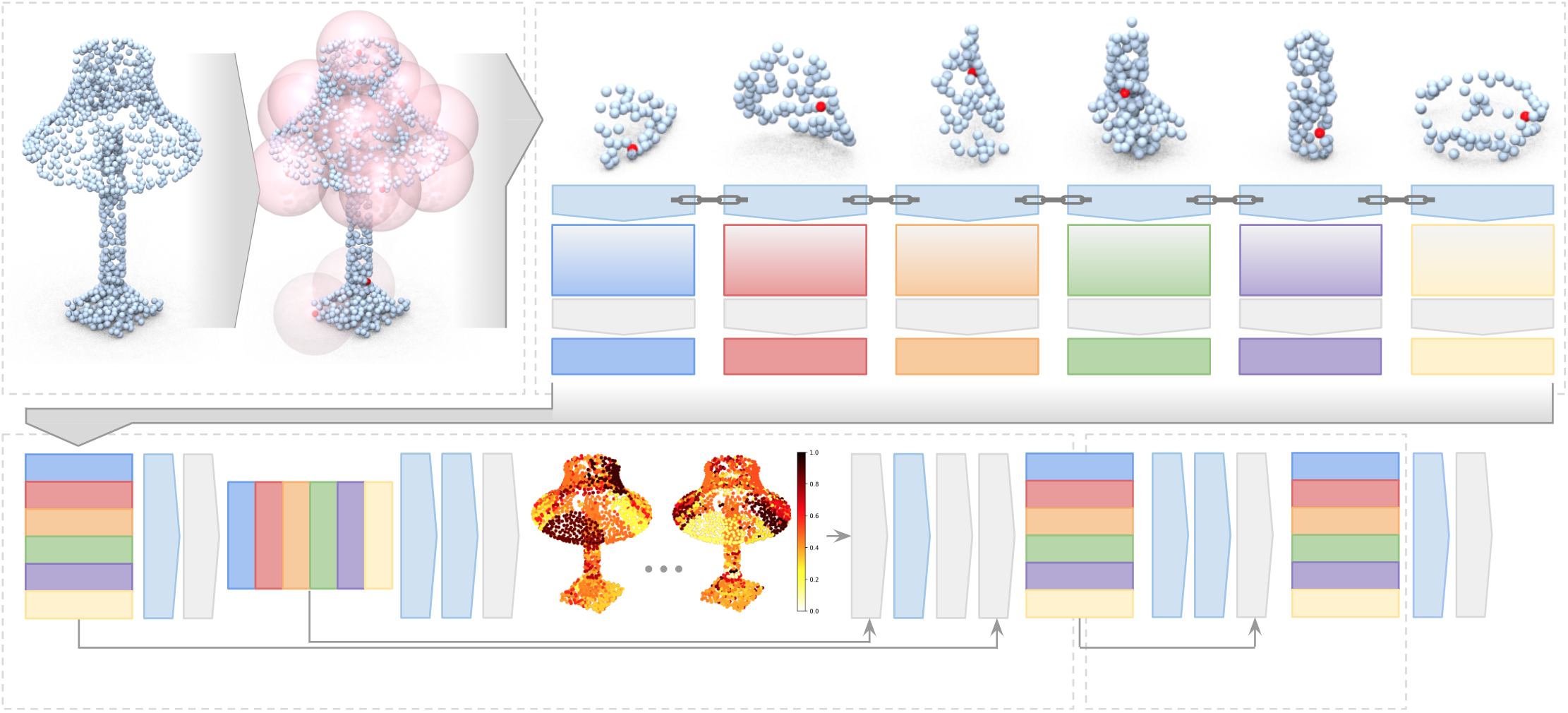}

        \put(3.5, 22){\small $\mathcal{X} \in \mathbb{R}^{V \times 3}$}
        \put(15.5, 22){\small $\{ \textbf{x}_i \in \mathbb{R}^3, i=1, \dots, P\}$}
        \put(14.33, 32){\small \rotatebox{90}{FPS}}
        \put(30.33, 29){\small \rotatebox{90}{Ball query}}

        \put(35, 41){\small $\mathcal{P}_{\textbf{x}_1, R} \in \mathbb{R}^{S \times 3}$}
        \put(37.5, 32){\small MLP$_1$}
        \put(36, 28){\small $\textbf{l}_1 \in \mathbb{R}^{S \times F}$}
        \put(36.5, 24.85){\small Pooling$_1$}
        \put(36.75, 22.15){\small $\textbf{p}_1 \in \mathbb{R}^{F}$}

        \put(88.5, 42){\small $\mathcal{P}_{\textbf{x}_P, R} \in \mathbb{R}^{S \times 3}$}
        \put(92.25, 32){\small MLP$_1$}
        \put(90.5, 28){\small $\textbf{l}_P \in \mathbb{R}^{S \times F}$}
        \put(91.25, 24.85){\small Pooling$_1$}
        \put(91.5, 22.15){\small $\textbf{p}_P \in \mathbb{R}^{F}$}

        \put(9.5, 7){\small \rotatebox{90}{LayerNorm}}
        \put(12.1, 7.5){\small \rotatebox{90}{Transpose}}

        \put(25.9, 6.75){\small \rotatebox{90}{Conv1d$_{P \times P}$}}
        \put(28.7, 9){\small \rotatebox{90}{BN1d}}
        \put(31.2, 8){\small \rotatebox{90}{Sigmoid}}

        \put(54.6, 8){\small \rotatebox{90}{Multiply}}
        \put(57.4, 9){\small \rotatebox{90}{MLP$_2$}}
        \put(60, 7.5){\small \rotatebox{90}{Transpose}}
        \put(62.9, 9.5){\small \rotatebox{90}{Sum}}

        \put(73.7, 7){\small \rotatebox{90}{LayerNorm}}
        \put(76.5, 9){\small \rotatebox{90}{MLP$_3$}}
        \put(79.2, 9.5){\small \rotatebox{90}{Sum}}

        \put(90.3, 7){\small \rotatebox{90}{LayerNorm}}
        \put(93.2, 7.75){\small \rotatebox{90}{Pooling$_2$}}
        \put(96, 9){\small \rotatebox{90}{$\textbf{g} \in \mathbb{R}^{F}$}}

        \put(2, 2.1){\small $\textbf{P} \in \mathbb{R}^{P \times F}$}
        \put(15.5, 2.25){\small $\textbf{P}^\top \in \mathbb{R}^{F \times P}$}
        \put(33.5, 2.25){\small $\{ \textbf{a}_i \in [0, 1]^{P}, i=1, \dots, P \}$}
        \put(65, 2.25){\small $\textbf{H} \in \mathbb{R}^{P \times F}$}
        \put(82, 2.25){\small $\textbf{G} \in \mathbb{R}^{P \times F}$}

        \put(1, 43.4){\small \textcolor{gray!40}{Patch extraction}}
        \put(35, 43.4){\small \textcolor{gray!40}{Patch embedding}}
        \put(1, 0.75){\small \textcolor{gray!40}{Attentive token Mixer}}
        \put(77.5, 0.75){\small \textcolor{gray!40}{Channel Mixer}}

    \end{overpic}
    \caption{
        Block diagram of the backbone $\Phi_{\boldsymbol{\Theta}}$ architecture.
        Parametric layers are represented with light blue blocks, non-parametric ones are represented with grey blocks or arrows.
        $V$ denotes the number of vertices, $P$ the number of patches, $S$ the number of samples, and $F$ the feature dimension.
        Vertex features for each sample in a patch are denoted as $\textbf{L} = \{ \textbf{l}_1, \dots, \textbf{l}_P \} \in \mathbb{R}^{P \times S \times F}$.
        Patch features are denoted by $\textbf{P} = \{ \textbf{p}_1, \dots, \textbf{p}_P \} \in \mathbb{R}^{P \times F}$.
        The attention maps learned by the attentive token mixer are denoted as $\textbf{a}_i \in \mathbb{R}^P, i=1, \dots, P$.
        The hidden features computed by the attentive token mixer and provided in input to the channel mixer are denoted as $\textbf{H} \in \mathbb{R}^{F \times P}$.
        The global feature for the whole shape is denoted as $\textbf{g}\in \mathbb{R}^{F}$.
        MLP$_i$, $i=1, 2, 3$, denotes the different MLP blocks.
        MLP$_1$ is shared across patches and has a pyramidal structure, i.e. the output feature dimension $F$ is higher than the input one.
        MLP$_{2,3}$ has a isotropic structure: the output feature dimension $F$ is the same as the input dimension.
        The visualization shows the simplified case where only a single shape is processed, however in practice a mini-batch of $B$ shapes are processed in parallel.
    }
    \label{fig:block_diagram}
\end{figure*}
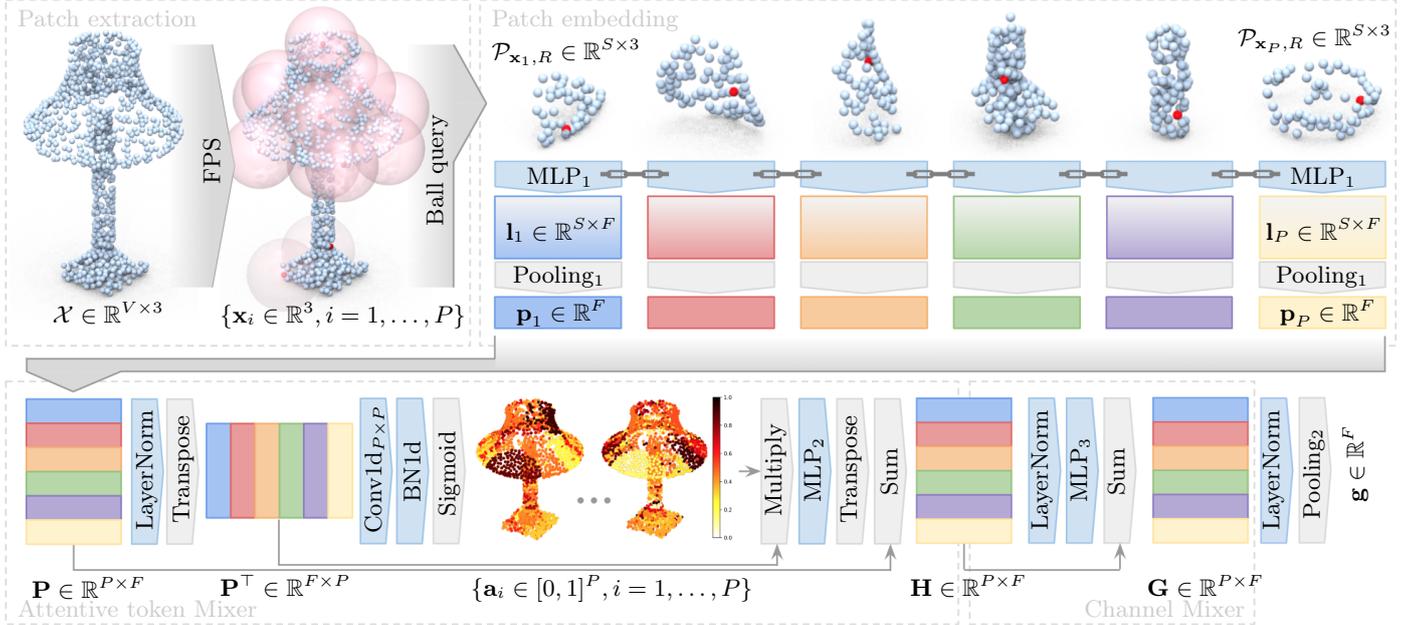

Our model consists of a backbone $\Phi_{\boldsymbol{\Theta}}$ and a task-specific head $\Psi_{\boldsymbol{\Omega}}$, where $\Phi, \Psi$ are parametric functions with learnable parameters $\boldsymbol{\Theta}, \boldsymbol{\Omega}$, respectively.
The backbone $\Phi_{\boldsymbol{\Theta}}$ is a feature extractor: it takes as input a point cloud $\mathcal{X} = \{ \textbf{x}_i \in \mathbb{R}^3, i=1, \dots, V \}$ of $V$ vertices and produces as output both local and global features capturing geometric information at different levels of detail.
The design of the head $\Psi_{\boldsymbol{\Omega}}$ depends on the downstream task:
classification uses global features only, while part segmentation uses both global and local features.

\subsection{Backbone architecture}
The backbone $\Phi_{\boldsymbol{\Theta}}$ consists of five modules: patch extraction, patch embedding, attentive token mixer, channel mixer, and feature aggregation. 
Fig.~\ref{fig:block_diagram} shows the backbone block diagram.

\textit{Patch extraction} 
takes as input the point cloud $\mathcal{X}$ and produces as output a set of $P$ local patches $\{ \mathcal{P}_i, i=1, \dots, P \}$, as shown in the top left corner of Fig.~\ref{fig:block_diagram}.
To extract such patches, we need a set of centroids and a radius $R$.
We use the Farthest Point Sampling (FPS) algorithm~\cite{PN2} to find $P$ evenly spread centroids out of the original $V$ vertices of $\mathcal{X}$.
For each centroid $\textbf{x}_i$, we extract a local patch of radius $R$ around it by using a ball query operation~\cite{PN2} as
\begin{equation}
 \mathcal{P}_i = \mathcal{P}_{\textbf{x}_i, R} = \{ \textbf{x}_j \in \mathcal{X} : \lVert \textbf{x}_j - \textbf{x}_i \rVert \le R \},
\end{equation}
where $\lVert \cdot \rVert$ denotes the Euclidean norm.
Typically, vertex density in a point cloud is not uniform: it may depend on both the level of geometric details and the acquisition viewpoint.
Hence, patches centered around different centroids may have different cardinalities.
To cope with this problem, we randomly sample with replacement $S$ points within each patch.
This allows us to 
i) create mini-batches of patches with the same cardinality that can be processed efficiently in parallel,
ii) mitigate overfitting by sampling different vertices at each training epoch, and
iii) create patches with more uniform vertex densities, enabling robustness to different densities of the input shape.

\textit{Patch embedding} 
takes as input the patches $\{ \mathcal{P}_i, i=1, \dots, P \}$ and learns how to embed them in a $F$-dimensional feature space, as shown in the top-right corner of Fig.~\ref{fig:block_diagram}.
The patch embedding design consists of positional encoding, feature encoding, and feature aggregation.
Positional encoding is implemented as a non-parametric layer taking as input the 3-dimensional absolute coordinates of each patch vertex $\textbf{x}_j \in \mathcal{P}_i, j=1, \dots, S$, and producing an enriched representation of its position by concatenating:
i) its relative coordinates with respect to the patch centroid $\textbf{x}_j - \textbf{x}_i$,
ii) its Euclidean distance from the patch centroid $\lVert \textbf{x}_j - \textbf{x}_i \rVert$, and
iii) the absolute coordinates of the patch centroid $\textbf{x}_i$.
Feature encoding is implemented as a MLP (MLP$_1$ in Fig.~\ref{fig:block_diagram}) taking as input the positional encoding and producing $F$-dimensional features $\textbf{l}_i \in \mathbb{R}^{S \times F}$ for each vertex of the patch $\mathcal{P}_i, i=1, \dots, P$.
The learnable weights of the MLP layer are shared across each of the $P$ patches to keep the number of parameters under control.
The feature aggregation is implemented as a max pooling operation along the patch points to aggregate local features $\textbf{l}_i \in \mathbb{R}^{S \times F}$ in a patch-wise global feature, namely token, $\textbf{p}_i \in \mathbb{R}^{F}$ describing the whole patch,
\begin{equation}
\textbf{p}_i = \max_{s = 1, \dots, S} \textbf{l}_i (s, \cdot), \quad i = 1, \dots, P.
\end{equation}
The output of the patch embedding module is a tensor $\textbf{P} \in \mathbb{R}^{P \times F}$ obtained by stacking the tokens $\textbf{p}_i \in \mathbb{R}^F$ for $i = 1, \dots, P$.

\textit{Attentive token mixer}
takes as input the patch descriptors $\textbf{P} \in \mathbb{R}^{P \times F}$ and produces as output a tensor of hidden features $\textbf{H} \in \mathbb{R}^{P \times F}$ where tokens are mixed together (Fig.~\ref{fig:block_diagram}, bottom left).
Feature mixing has proven to be critical for the design of an effective deep architecture.
Competitor models designs consider different strategies to merge point-wise features:
PointNet++~\cite{PN2} concatenates vertex features from different levels of a hierarchy,
DGCNN~\cite{DGCNN} learns filters to aggregate vertex features connected by local graphs,
while PointTransformer~\cite{PointTransformer} merges vertex features in a local neighborhood by means of a self-attention mechanism.
Instead, the attentive token mixer learns how to aggregate patch-wise features representing shape parts instead of individual vertices.
It is implemented as a MLP (MLP$_2$ in Fig.~\ref{fig:block_diagram}) trained contextually to the feature extractor.

The vanilla version of the token mixer can be implemented as a residual block with two linear layers separated by an activation function $\sigma$, such as
\begin{equation}\label{eq:token_mixer}
\textbf{H} = \text{Lin}_2( \sigma ( \text{Lin}_1 ( \textbf{P} ) ) ) + \textbf{P},
\end{equation}
where $\text{Lin}_1$ takes as input features of size $F$ and produces as output features of size $F_\text{T}$, while $\text{Lin}_2$ does the opposite.
This symmetric design allows to add a residual connection to the input features which has proven to simplify the network's optimization by facilitating the gradient propagation and by smoothing the loss landscape~\cite{ResNet}.

However, mixing information across all tokens can be detrimental to the performance of the network since different tokens may contain semantically inconsistent information.
For this reason, we introduce a learnable attention mechanism to select which tokens should be mixed with others in order to create a global shape descriptor robust to missing parts and local variations in style and shape.
This can be achieved by modifying Eq.~\ref{eq:token_mixer} as
\begin{equation}\label{eq:attentive_token_mixer}
\textbf{H} = \text{Lin}_2 \left( \text{Att}_2 ( \tilde{\textbf{H}} ) \odot \tilde{\textbf{H}} \right) + \textbf{P},
\end{equation}
where
\begin{equation}
\tilde{\textbf{H}} = \sigma \left( \text{Lin}_1 \left( \text{Att}_1 ( \textbf{P} ) \odot \textbf{P} \right) \right),
\end{equation}
$\text{Att}_{1,2}(\cdot) = \text{sigmoid}( \text{BN1d}( \text{Conv1d}(\cdot) ) )$, and $\odot$ denotes batch-wise matrix multiplication.
Thanks to the sigmoid activation function, the attention modules produces as output a probability distribution $\textbf{a}_i \in [0, 1]^P$ over the patches $\mathcal{P}_i, i=1,\dots,P$.
Please note that compared to the self-attention mechanism used by PointTransformer~\cite{PointTransformer}, which acts on vertex features locally, the proposed attentive token mixer is fundamentally different: it computes the attention across patch-wise features encoding shape parts, and works at a global scale allowing the merging of information across potentially distant shape parts.

\textit{Channel mixer}
takes as input the tensor $\textbf{H} \in \mathbb{R}^{P \times F}$ and produces as output a tensor $\textbf{G} \in \mathbb{R}^{P \times F}$ where the channels of the tokens are mixed (Fig.~\ref{fig:block_diagram}, bottom right corner).
The channel mixer is implemented as Eq.~\ref{eq:token_mixer} by replacing $\textbf{P}, \textbf{H}$ with $\textbf{H}, \textbf{G}$, respectively.
The interaction between patch feature channels allows the channel-mixing MLP (MLP$_3$ in Fig.~\ref{fig:block_diagram}) to learn more discriminative embeddings.
In contrast to the PointNet~\cite{PN}, PointNet++~\cite{PN2}, and DGCNN~\cite{DGCNN} architectures where the MLP blocks expand the input features dimension, the channel mixer has a symmetric design where output features has the same size of the input ones.
The symmetric design of the attentive token mixer and channel mixer modules allows to add a skip connection between $\textbf{G}$ and $\textbf{P}$.
This means that we can iterate through the mixer modules to increase the depth of the model.
The number of such iterations $D$ is an hyper-parameter of the model.

\textit{Feature aggregation}
takes as input the patch features $\textbf{G} \in \mathbb{R}^{P \times F}$ and aggregates them using max pooling to create a global feature $\textbf{g} \in \mathbb{R}^F$ encoding the whole shape $\mathcal{X}$, i.e.
\begin{equation}
\textbf{g} = \max_{p = 1, \dots, P} \textbf{G} (p, \cdot).    
\end{equation}
The global feature is then passed to the task-specific head.

\subsection{Design of task-specific heads}

The \textit{classification head}
takes as input the global feature $\textbf{g} \in \mathbb{R}^F$ and produces as output a probability distribution over the $C$ object categories present in the dataset at hand, i.e. $\textbf{c} = \Psi_{\boldsymbol{\Omega}}(\textbf{g}) \in [0, 1]^C$.
$\Psi_{\boldsymbol{\Omega}}$ is implemented as an MLP by stacking three blocks containing a linear layer, batch normalization, an activation function, and dropout.
During training, a cross-entropy loss function will force the logits produced by $\Psi_{\boldsymbol{\Omega}}$ to match the target annotations.

The \textit{part segmentation head} requires a modification of the backbone architecture to generate $F$-dimensional features for each vertex of each patch~\cite{PN}.
The part segmentation head $\Psi_{\boldsymbol{\Omega}}$ takes as input local features and produces as output a probability distribution of size $PS \times C$, where $C$ is the number of shape parts.
$\Psi_{\boldsymbol{\Omega}}$ is implemented as an MLP with four blocks.
The first block reduces the feature dimension from $3F$ to $F$, the other three blocks are the same of the classification head.
The global feature $\textbf{g} \in \mathbb{R}^F$ is repeated $P$ times and concatenated with the patch features $\textbf{P} \in \mathbb{R}^{P \times F}$ to obtain a tensor of size $P \times 2F$.
Such tensor is then repeated $S$ times and concatenated with the local features $\textbf{L} \in \mathbb{R}^{P \times S \times F}$ to obtain a tensor of size $P \times S \times 3F$, which is reshaped to size $PS \times 3F$.
This design is to enrich the feature representation by incorporating both local geometry and global semantics.
Part segmentation benchmarks usually provide a label for each of the $V$ vertices of $\mathcal{X}$.
However, the output of $\Psi_{\boldsymbol{\Omega}}$ is structured differently and requires a target label for each vertex of each patch.
We can extract the desired ground-truth annotation from the one available at shape-level by resorting to the correspondences between points on the shape and points on the patches.
This correspondence is provided by the ball query operation done in the patch extraction module.
Therefore, we can create a ground-truth supervision tensor $\textbf{t} \in [1, \dots, C]^{PS}$.
During training, a cross-entropy loss between the output of $\Psi_{\boldsymbol{\Omega}}$ and the target annotation $\textbf{t}$ is minimized.

\section{Experiments}\label{sec:experiments}

\subsection{Comparison setup}
We compare PatchMixer with a selection of the most common deep architectures: PointNet~\cite{PN}, PointNet++~\cite{PN2}, DGCNN~\cite{DGCNN}, PointTransformer~\cite{PointTransformer}, PointMLP~\cite{PointMLP}, and PointMixer~\cite{PointMixer}.
All the experiments are run by us by using the same training settings.

\subsection{Classification}\label{sec:classification}
The goal of shape classification is to predict the category to which the observed point cloud belongs to, given a predefined set of classes. 

\subsubsection{Datasets}
We use GraspNetPC~\cite{GAI} and PointDA~\cite{PointDAN} datasets and they include different and complementary challenges.

GraspNetPC is composed of four subsets:
shapes acquired in a real-world environment with a Kinect2 sensor (RK) and with a Intel RealSense sensor (RR),
and shapes created synthetically with computer graphics software by emulating both sensors settings (SK, SR).
It contains point clouds from ten categories: box, can, banana, powerdrill, scissors, pear, dish, camel, mouse, and shampoo.
GraspNetPC is particularly suited to test the robustness of a classifier to partiality and acquisition noise since point clouds acquired with depth sensors are typically affected by self-occlusions and clutter.
Synthetic and real subsets of GraspNetPC have similar intra-class variability, because the synthetic point clouds were created to emulate the real-world ones.

PointDA is composed of point cloud of furniture objects from ten common categories (bathtub, bed, bookshelf, cabinet, chair, lamp, monitor, plant, sofa, and table) from three widely-used datasets: ModelNet~\cite{Wu2015}, ShapeNet~\cite{ShapeNet}, and ScanNet~\cite{ScanNet}.
Point clouds of ModelNet and ShapeNet are obtained by sampling synthetic 3D CAD models from two different data distributions: ShapeNet objects are gathered from online repositories and have larger variance in structure compared to ModelNet.
Point clouds of ScanNet are extracted by real-world indoor scenes acquired with RGBD sensors, thus are occluded by surroundings and may lose some of their parts according to the acquisition viewpoint.
Unlike GraspNetPC, the variability of object poses in PointDA is limited because its point clouds are aligned so that their maximum elongation is along the gravity axis.

\subsubsection{Network configuration}
All the experiments denoted as ``Ours'' are performed with the same network configuration.
We randomly sample the input shape to $V=1024$ vertices.
We extract $P=256$ patches.
Each patch is sampled to $S=128$ vertices using random sampling with replacement.
To improve the robustness against missing parts we use patch masking to drop around 30\% of the patches before feeding them to the patch embedding.
Patch embedding produces features of dimension $F=1024$.
The patch embedding consists of an MLP stacking five Conv1d + BN1d + ReLU blocks.
The output channels dimensions of the Conv1d layers are 16, 32, 64, 128, and 1024.
The hidden feature dimensions for the mixer layers are set to $F_\text{T}=F_\text{C}=512$.
The depth of the mixer modules is set to $D=1$.

\begin{table}[t!]
    \centering
    \tabcolsep 3pt
    \caption[]{
        Classification performance on GraspNetPC~\cite{GAI} evaluated in terms of OA.
        Keys: RR: Real RealSense, RK: Real Kinect, SR: Synthetic RealSense, SK: Synthetic Kinect, TL: transfer learning.
        The X$\to$Y columns report performance obtained when training on the training split of dataset X and testing on the test split of dataset Y.
        Gray background denotes the same domain scenario.
        The AvgTL column reports the average OA performance in the TL scenario.
    }
    \label{tab:graspnetpc}
    \smallskip
    \resizebox{1.0\columnwidth}{!}{%
        \begin{tabular}{lgcccc}
            \toprule
            Method & RR$\to$RR & RK$\to$RR & SR$\to$RR & SK$\to$RR & AvgTL \\
            \toprule
            PointNet~\cite{PN} & 0.869 & 0.754 & 0.670 & 0.742 & 0.732 \\
            PointNet++ MSG~\cite{PN2} & 0.947 & 0.496 & 0.616 & 0.639 & 0.584 \\
            DGCNN~\cite{DGCNN} & 0.941 & 0.627 & 0.681 & 0.724 & 0.677 \\
            PointTransformer~\cite{PointTransformer} & 0.940 & 0.632 & 0.694 & 0.742 & 0.689 \\
            PointMLP~\cite{PointMLP} & 0.971 & 0.688 & 0.720 & 0.798 & 0.735 \\
            PointMixer~\cite{PointMixer} & 0.944 & 0.690 & 0.728 & 0.815 & \underline{0.744} \\
            Ours & 0.950 & 0.782 & 0.786 & 0.832 & \textbf{0.800} \\
            \toprule
            Method & RK$\to$RK & RR$\to$RK & SK$\to$RK & SR$\to$RK & AvgTL \\
            \toprule
            PointNet~\cite{PN} & 0.969 & 0.792 & 0.802 & 0.653 & 0.749 \\
            PointNet++~\cite{PN2} & 0.979 & 0.822 & 0.787 & 0.732 & 0.780 \\
            DGCNN~\cite{DGCNN} & 0.989 & 0.820 & 0.906 & 0.810 & 0.846 \\
            PointTransformer~\cite{PointTransformer} & 0.991 & 0.898 & 0.891 & 0.870 & 0.886 \\
            PointMLP~\cite{PointMLP} & 0.997 & 0.914 & 0.934 & 0.901 & \underline{0.916} \\
            PointMixer~\cite{PointMixer} & 0.992 & 0.847 & 0.925 & 0.886 & 0.886 \\
            Ours & 0.985 & 0.936 & 0.938 & 0.902 & \textbf{0.925} \\
            \bottomrule
        \end{tabular}
    }
\end{table}

\subsubsection{Training parameters}
We train all the models for 200 epochs using mini-batches of size 32.
We use stochastic gradient descent with a Nesterov momentum of 0.9.
We use weight decay with a factor of $10^{-5}$.
The initial learning rate is set to $10^{-2}$, which is progressively reduced during training to $10^{-4}$ using a cosine annealing scheduler with a period equal to the number of epochs~\cite{CosineAnnealing}.
We use data augmentation to avoid overfitting and gain robustness to different object transformations and noise:
we first center the shape according to its barycenter and scale it to have unit diagonal, 
we jitter the position of each point by Gaussian noise with zero mean and $10^{-2}$ standard deviation,
we randomly rotate the object along the up-axis,
and finally scale it by a random factor in the interval (0.8, 1.2).

\begin{table}[t!]
    \centering
    \caption[]{
        Classification performance on PointDA~\cite{PointDAN} evaluated in terms of OA.
        Dataset keys: M:~ModelNet, S:~ShapeNet, $\text{S}^\prime$:~ScanNet.
        The structure follows Table~\ref{tab:graspnetpc}.
    }
    \label{tab:pointda}
    \smallskip
    \resizebox{1.0\columnwidth}{!}{%
        \begin{tabular}{lgccc}
            \toprule
            Method & M$\to$M & M$\to$S & M$\to$$\text{S}^\prime$ & AvgTL \\
            \toprule
            PointNet \cite{PN} & 0.966 & 0.834 & 0.475 & 0.655 \\
            PointNet++ SSG \cite{PN2} & 0.966 & 0.688 & 0.501 & 0.595 \\
            PointNet++ MSG \cite{PN2} & 0.980 & 0.755 & 0.511 & 0.633 \\
            DGCNN \cite{DGCNN} & 0.987 & 0.787 & 0.499 & 0.643 \\
            PointTransformer \cite{PointTransformer} & 0.973 & 0.764 & 0.339 & 0.552 \\
            PointMLP \cite{PointMLP} & 0.984 & 0.843 & 0.478 & \underline{0.660} \\
            PointMixer \cite{PointMixer} & 0.978 & 0.838 & 0.479 & 0.659 \\
            Ours & 0.971 & 0.831 & 0.514 & \textbf{0.673} \\
            \toprule
            Method & S$\to$S & S$\to$M & S$\to$$\text{S}^\prime$ & AvgTL \\
            \toprule
            PointNet \cite{PN} & 0.941 & 0.789 & 0.421 & 0.605 \\
            PointNet++ SSG \cite{PN2} & 0.942 & 0.775 & 0.441 & 0.608 \\
            PointNet++ MSG \cite{PN2} & 0.946 & 0.780 & 0.451 & 0.616 \\
            DGCNN \cite{DGCNN} & 0.948 & 0.776 & 0.443 & 0.610 \\
            PointTransformer \cite{PointTransformer} & 0.939 & 0.776 & 0.409 & 0.593 \\
            PointMLP \cite{PointMLP} & 0.948 & 0.800 & 0.459 & \underline{0.630} \\
            PointMixer \cite{PointMixer} & 0.945 & 0.822 & 0.422 & 0.622 \\
            Ours & 0.945 & 0.810 & 0.451 & \textbf{0.631} \\
            \toprule
            Method & $\text{S}^\prime \to \text{S}^\prime$ & $\text{S}^\prime \to \text{M}$ & $\text{S}^\prime \to \text{S}$ & AvgTL \\
            \toprule
            PointNet \cite{PN} & 0.761 & 0.649 & 0.666 & \textbf{0.657} \\
            PointNet++ SSG \cite{PN2} & 0.780 & 0.469 & 0.541 & 0.505 \\
            PointNet++ MSG \cite{PN2} & 0.788 & 0.526 & 0.586 & 0.556 \\
            DGCNN \cite{DGCNN} & 0.783 & 0.620 & 0.545 & 0.582 \\
            PointTransformer \cite{PointTransformer} & 0.772 & 0.582 & 0.580 & 0.581 \\
            PointMLP \cite{PointMLP} & 0.798 & 0.638 & 0.674 & \underline{0.656} \\
            PointMixer \cite{PointMixer} & 0.789 & 0.581 & 0.636 & 0.608 \\
            Ours & 0.795 & 0.641 & 0.671 & \underline{0.656} \\
            \bottomrule
        \end{tabular}
    }
\end{table}

\subsubsection{Quantitative results}

Table~\ref{tab:graspnetpc} shows the results on GraspNetPC~\cite{GAI} evaluated in term of Overall Accuracy (OA).
The top row shows the performance on the RR test split obtained with a model trained on the RR training split (same domain, gray background), and on the RK, SR, and SK training splits (transfer learning, white background).
The bottom row is structured in the same way but shows the performance obtained on the RK test split.
The last column shows the average OA performance in the Transfer Learning (TL) scenario (white background).
The best result is highlighted in bold, the second best result is underlined.

PatchMixer outperforms all the other backbone designs in the transfer learning scenario, while reaching performance very close to competitors on the same domain scenario.
On RR, PatchMixer outperforms PointMixer by 5.6\%, PointMLP by 6.5\%, PointTransformer by 11.1\%, DGCNN by 12.3\%, PointNet++ by 21.6\%, and PointNet by 6.8\%, in terms of AvgTL.
On RK, the gap against the second best method, PointMLP, reduces to around 1\%, while the difference with the other approaches is quite significant with a gap of more than 4\%.
On the same domain, PointMLP achieves the best performance at the cost of a much worse generalization: on RR$\to$RR PointMLP improves over PatchMixer by 2.1\% OA, but falls behind it by 9.4\% OA in RK$\to$RR.
This further consolidate our claim: a better performance in the same domain scenario is not sufficient to assess the quality of an architecture design.

\captionsetup{skip=0pt}
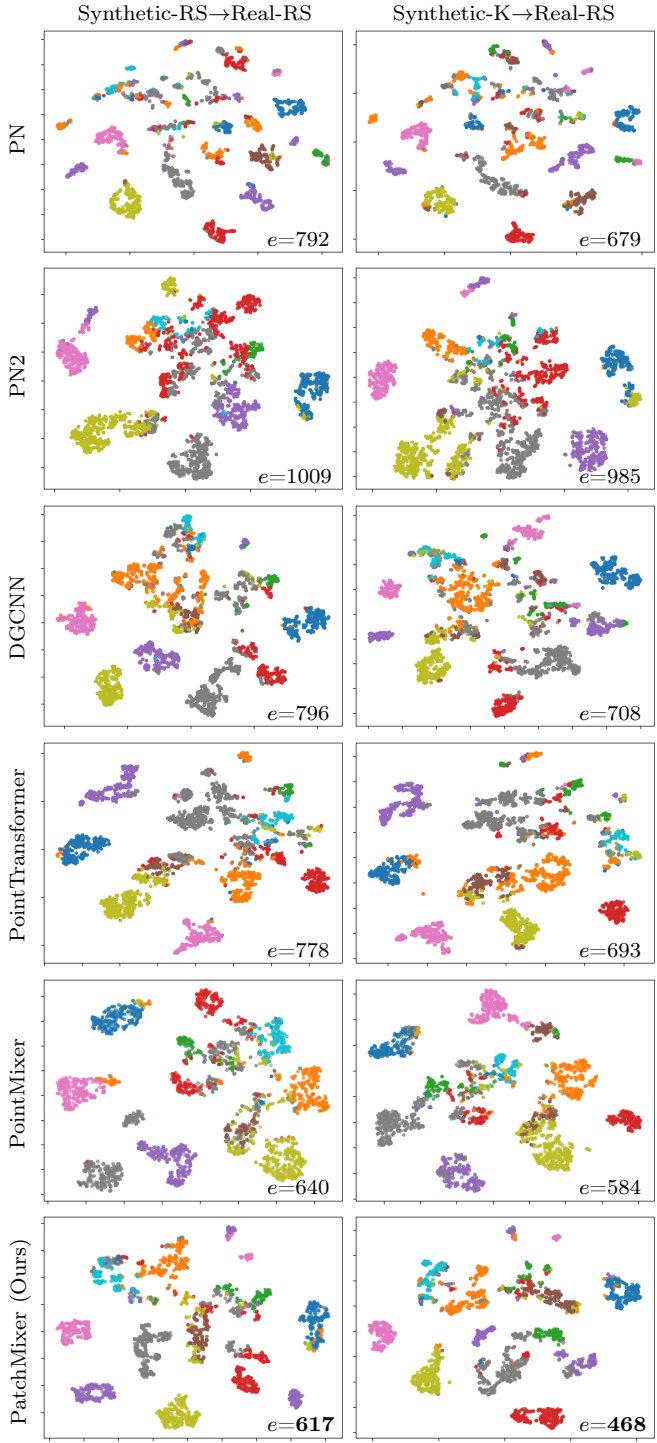
\begin{figure}[t!]
    \centering
    \begin{overpic}[width=0.45\columnwidth, trim=2mm 0 2mm 0, clip]{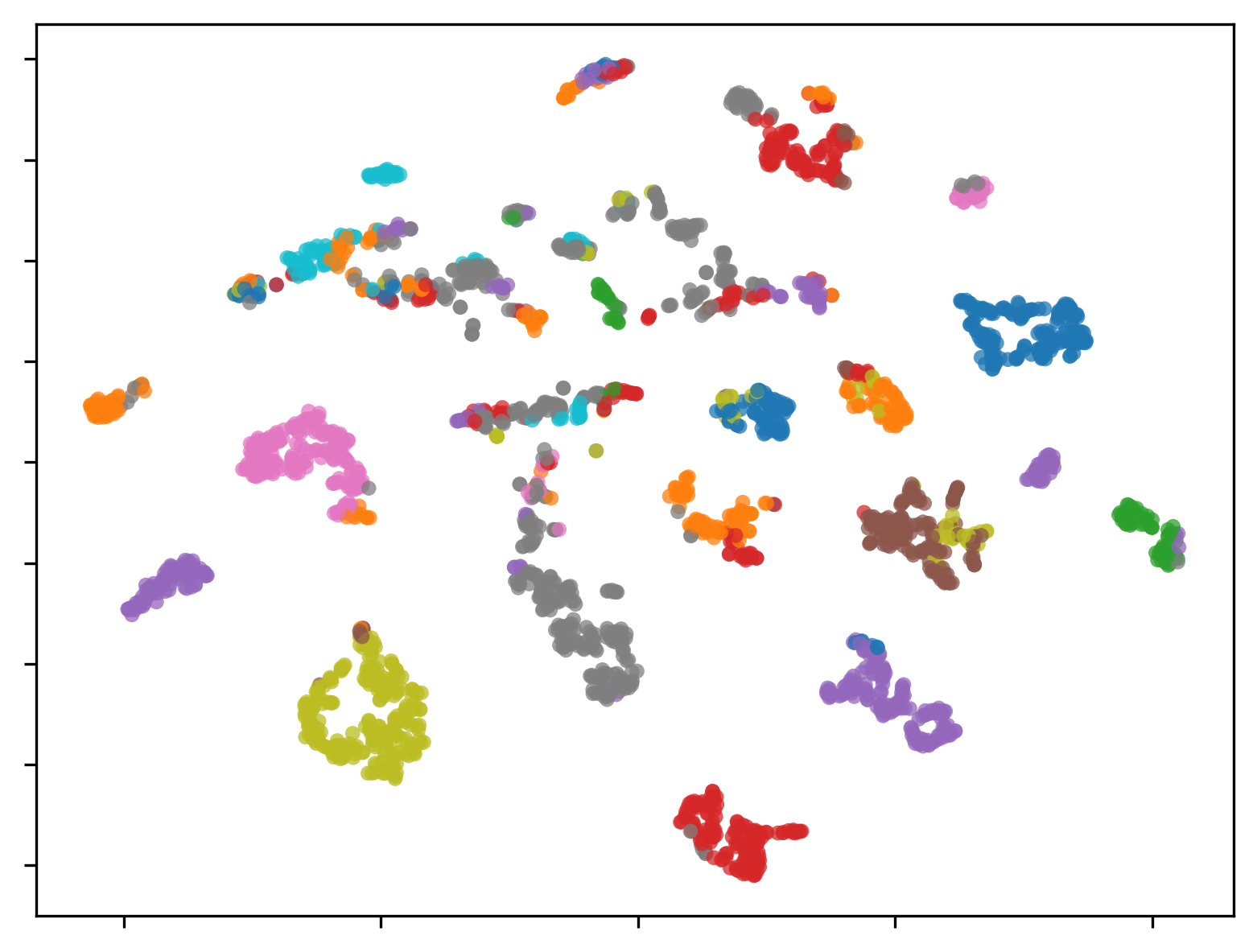}
        \put(12.5, 80){\footnotesize Synthetic-RS$\to$Real-RS}
        \put(-10, 35){\footnotesize \rotatebox{90}{PN}}
        \put(75, 5){\footnotesize $e$=792}
    \end{overpic}
    \begin{overpic}[width=0.45\columnwidth, trim=2mm 0 2mm 0, clip]{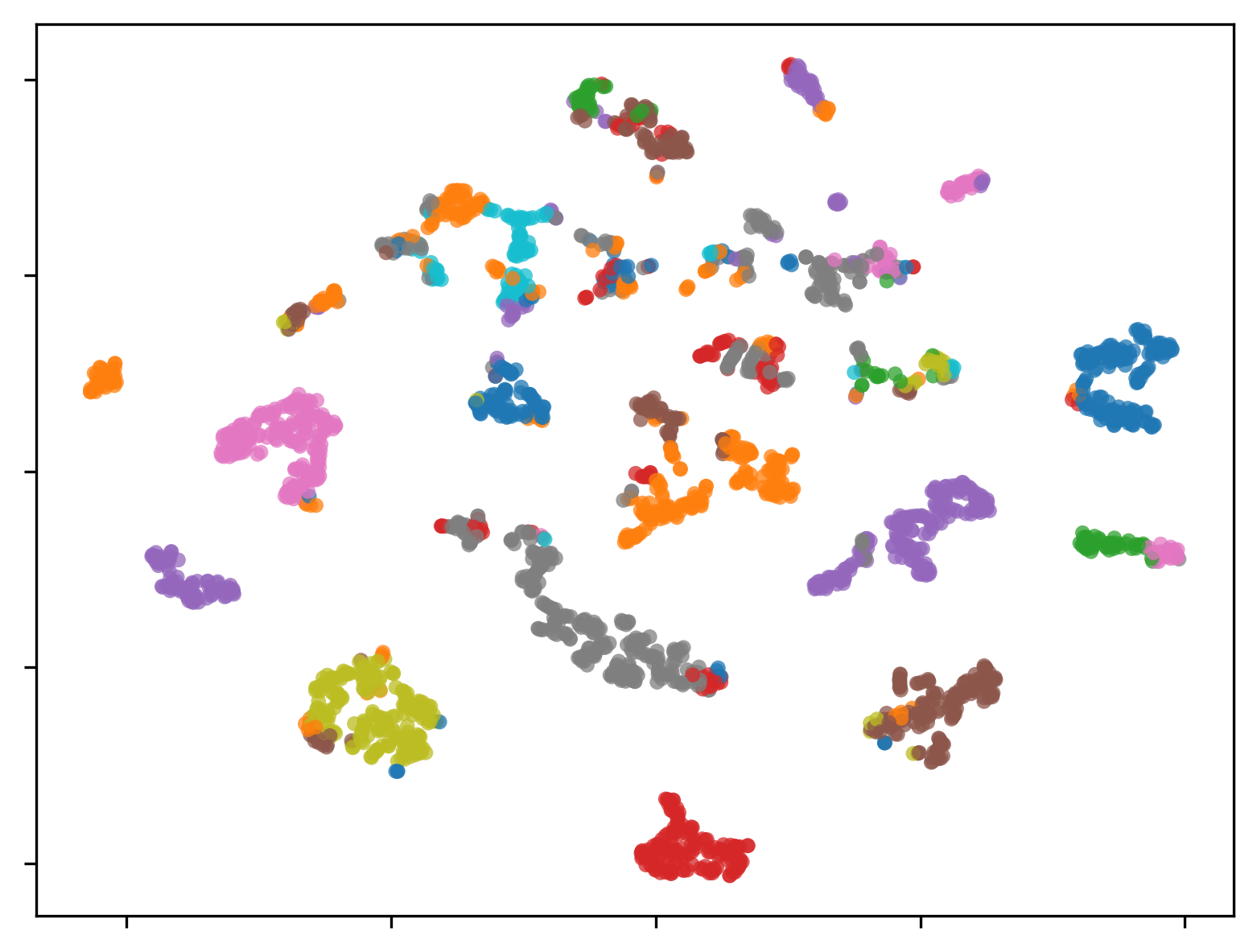}
        \put(13.5, 80){\footnotesize Synthetic-K$\to$Real-RS}
        \put(75, 5){\footnotesize $e$=679}
    \end{overpic}
    \\
    \begin{overpic}[width=0.45\columnwidth, trim=2mm 0 2mm 0, clip]{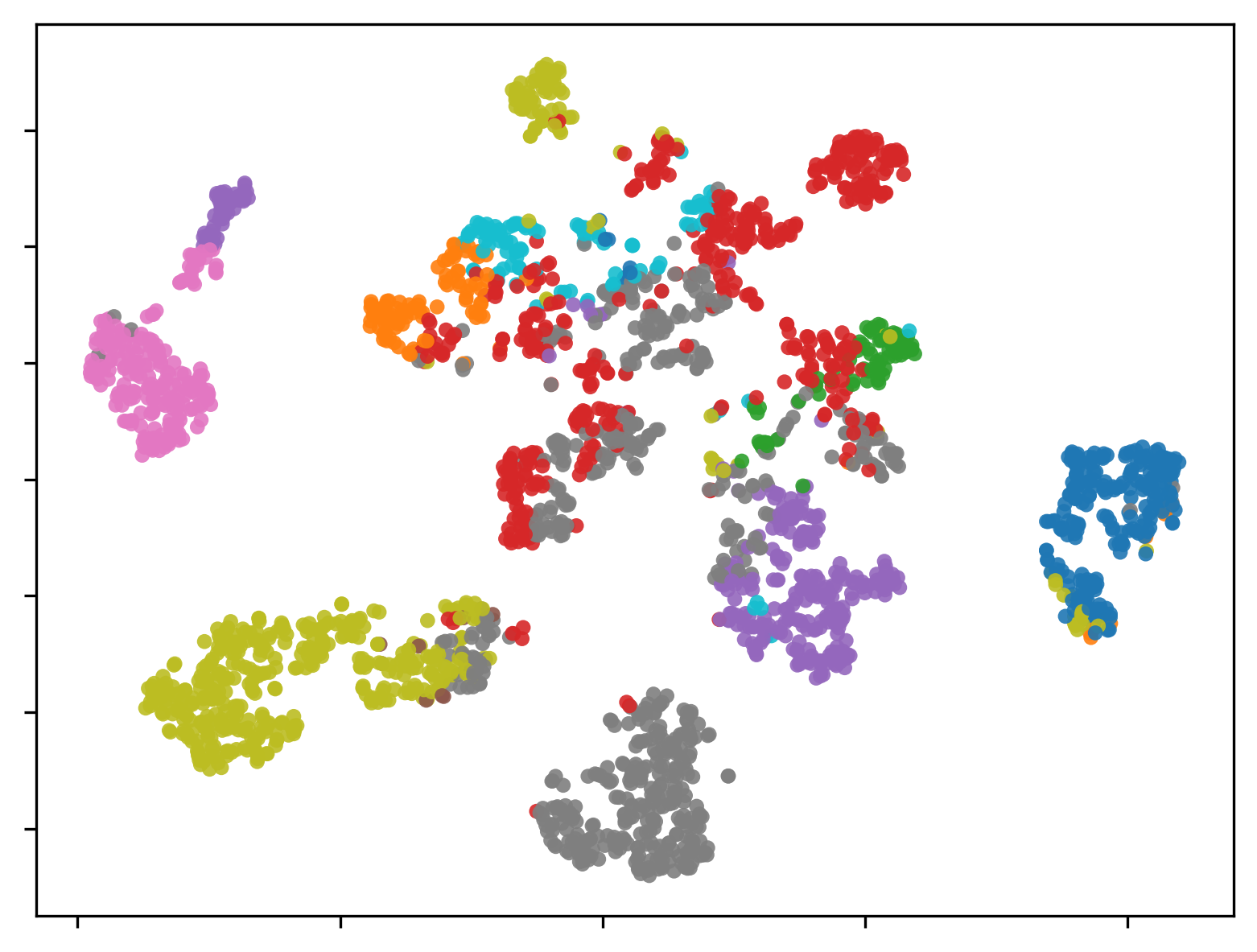}
        \put(-10, 33){\footnotesize \rotatebox{90}{PN2}}
        \put(72, 5){\footnotesize $e$=1009}
    \end{overpic}
    \begin{overpic}[width=0.45\columnwidth, trim=2mm 0 2mm 0, clip]{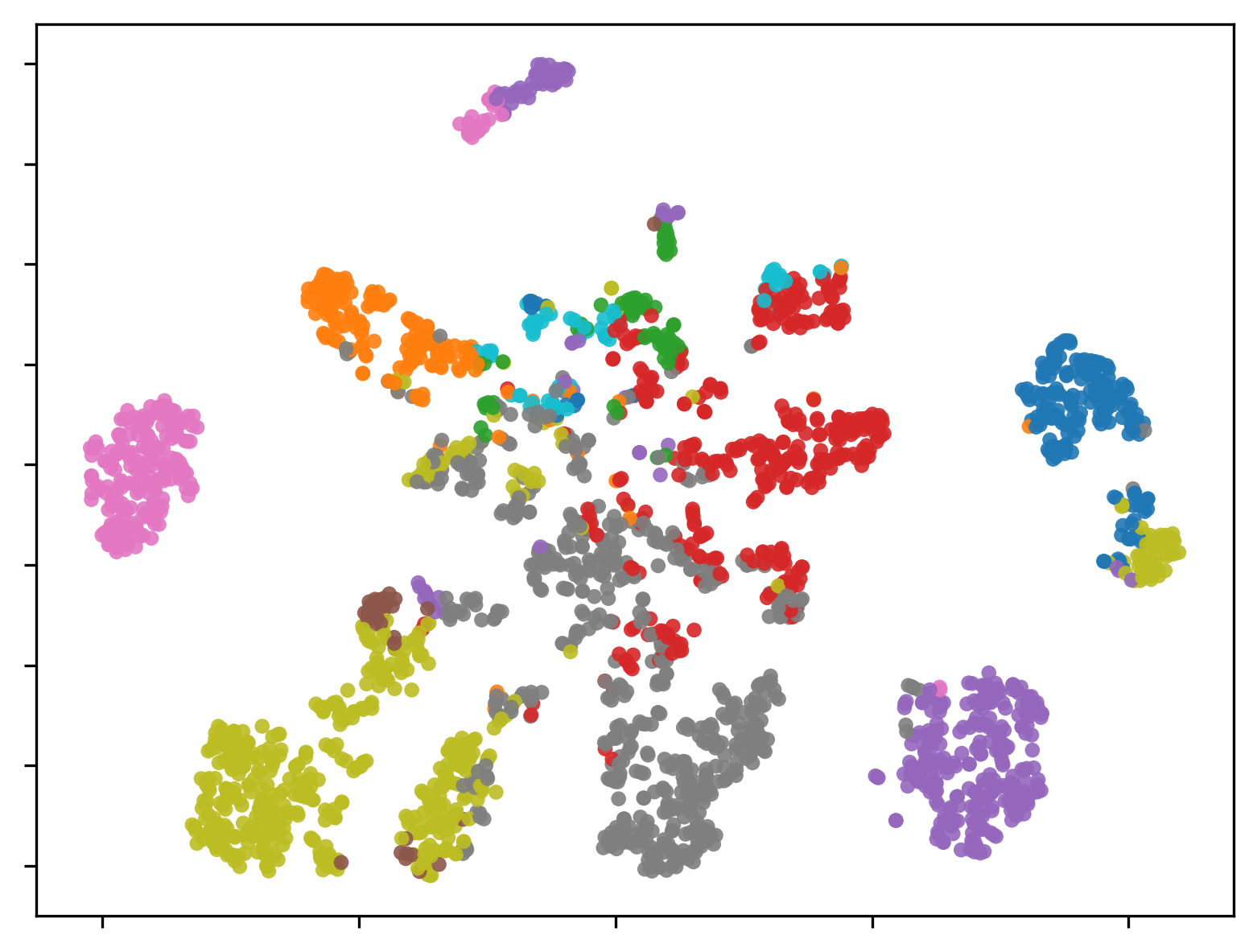}
        \put(75, 5){\footnotesize \contour{white}{$e$=985}}
    \end{overpic}
    \\
    \begin{overpic}[width=0.45\columnwidth, trim=2mm 0 2mm 0, clip]{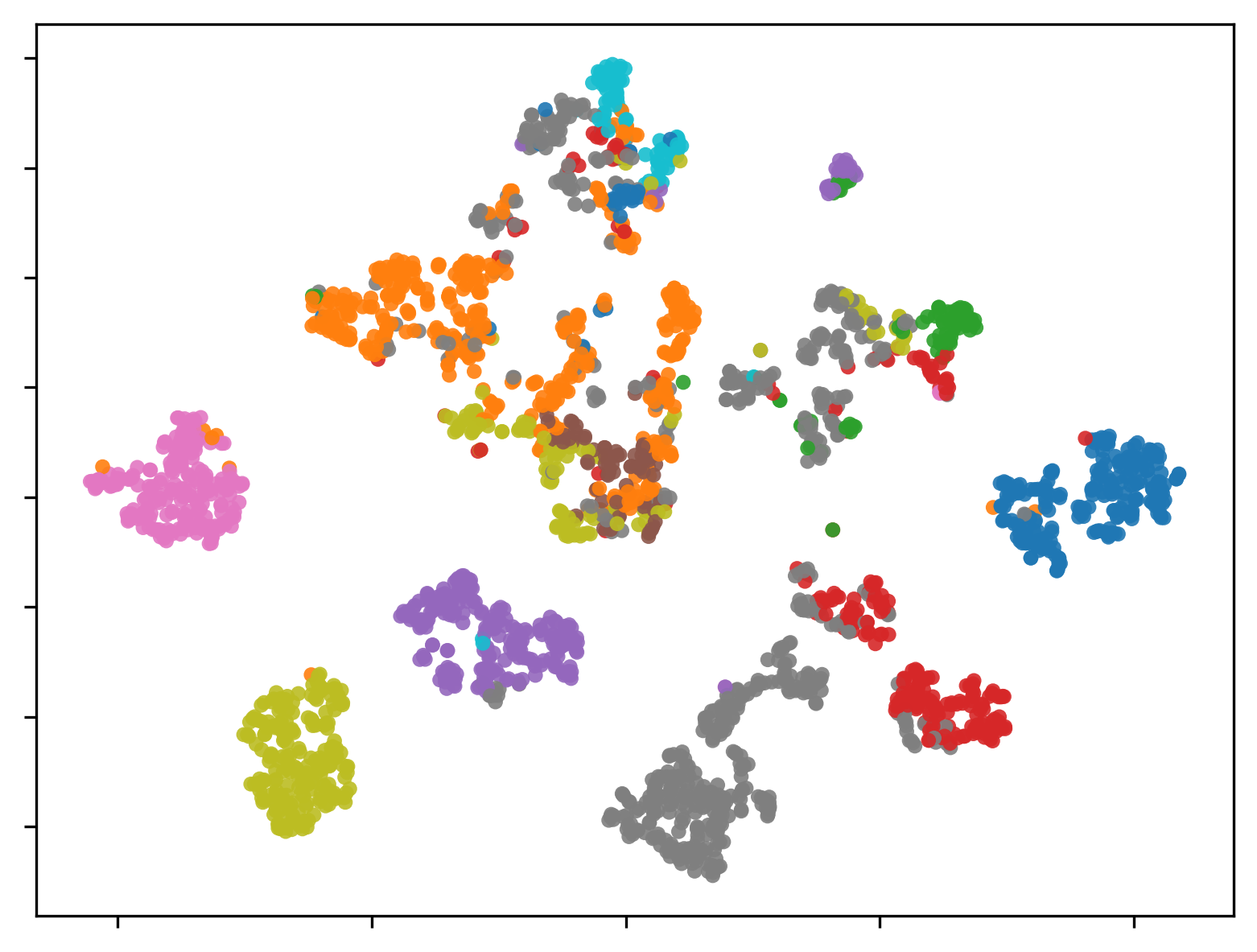}
        \put(-10, 25){\footnotesize \rotatebox{90}{DGCNN}}
        \put(75, 5){\footnotesize $e$=796}
    \end{overpic}
    \begin{overpic}[width=0.45\columnwidth, trim=2mm 0 2mm 0, clip]{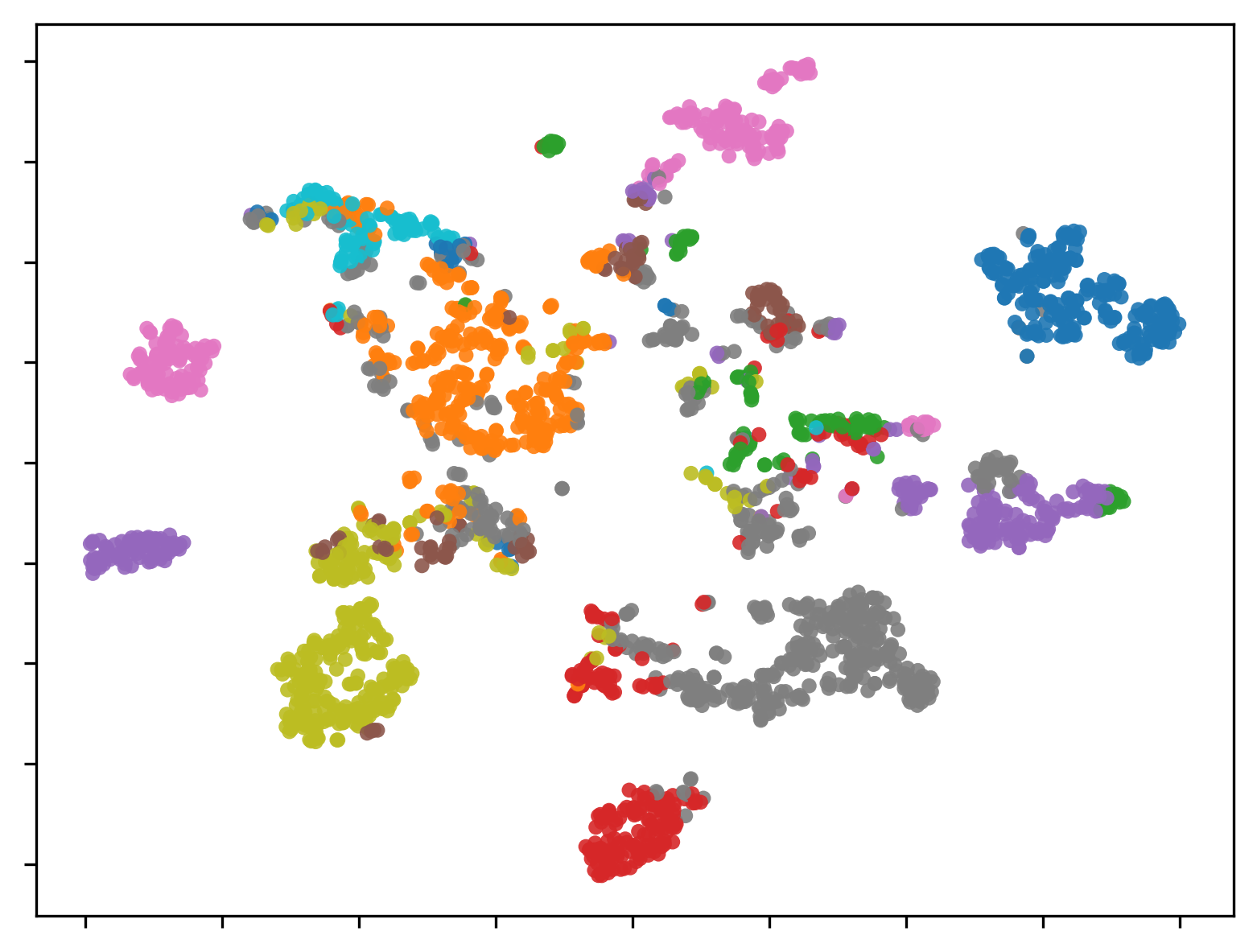}
        \put(75, 5){\footnotesize $e$=708}
    \end{overpic}
    \\
    \begin{overpic}[width=0.45\columnwidth, trim=2mm 0 2mm 0, clip]{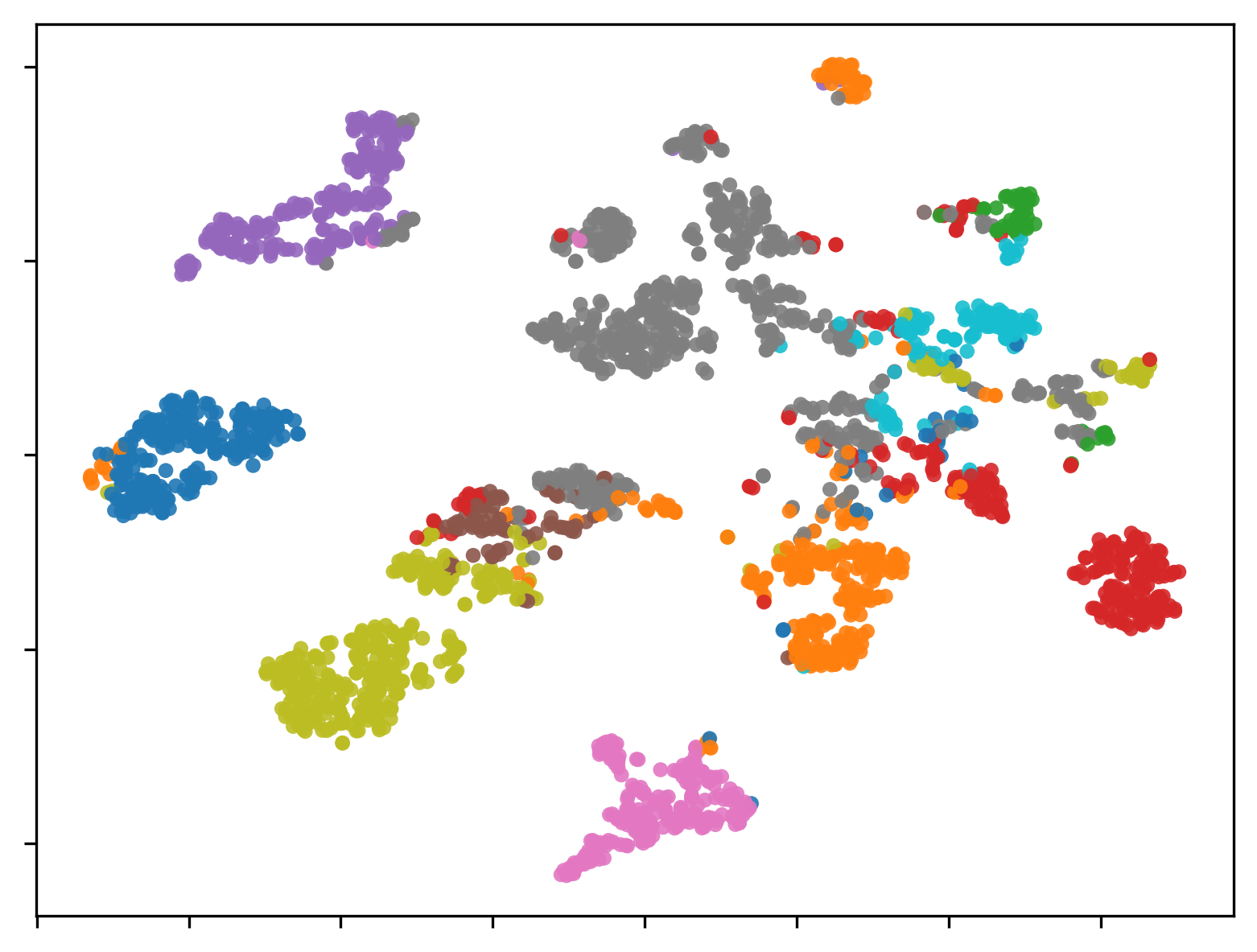}
        \put(-10, 10){\footnotesize \rotatebox{90}{PointTransformer}}
        \put(75, 5){\footnotesize $e$=778}
    \end{overpic}
    \begin{overpic}[width=0.45\columnwidth, trim=2mm 0 2mm 0, clip]{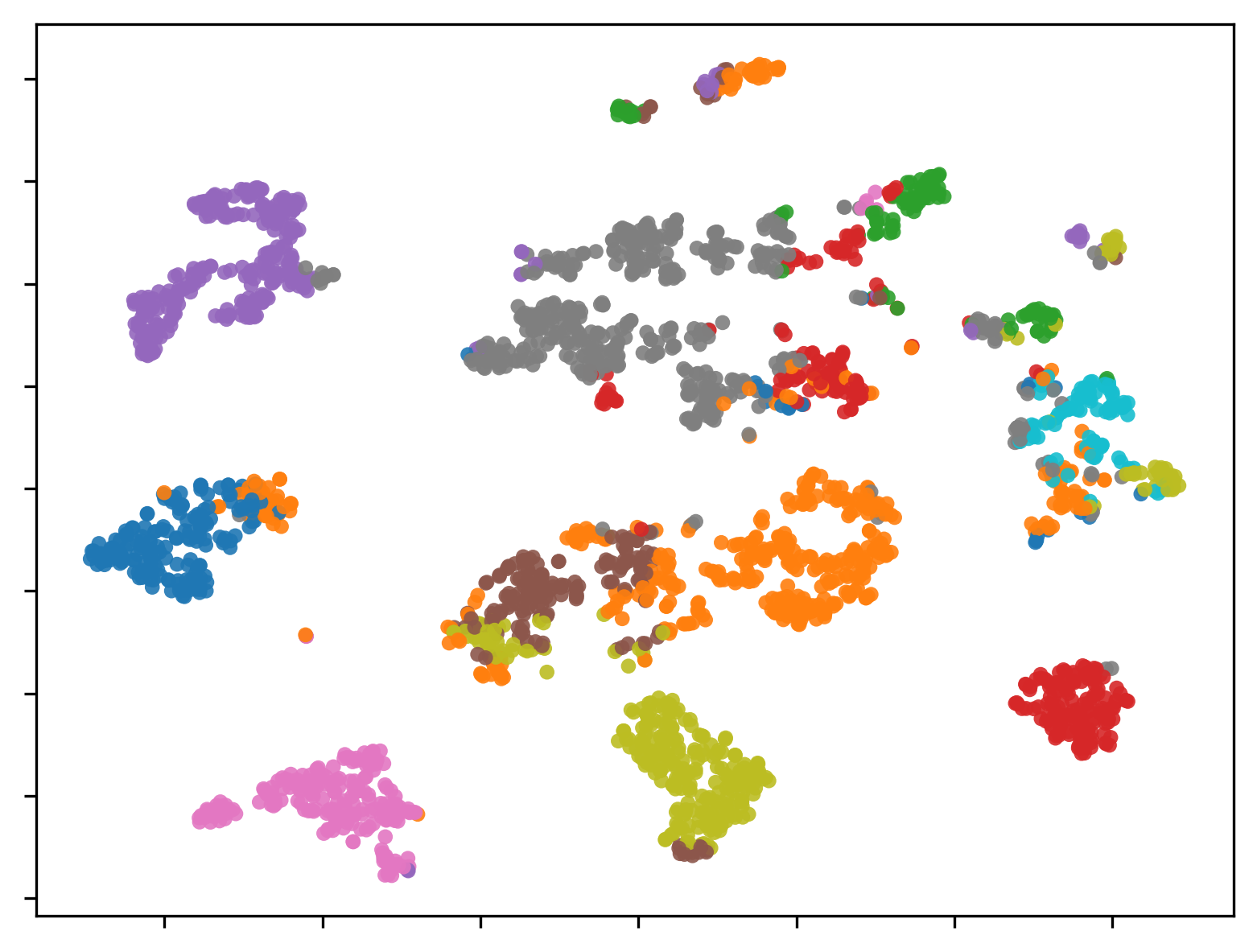}
        \put(75, 5){\footnotesize $e$=693}
    \end{overpic}
    \\
    \begin{overpic}[width=0.45\columnwidth, trim=2mm 0 2mm 0, clip]{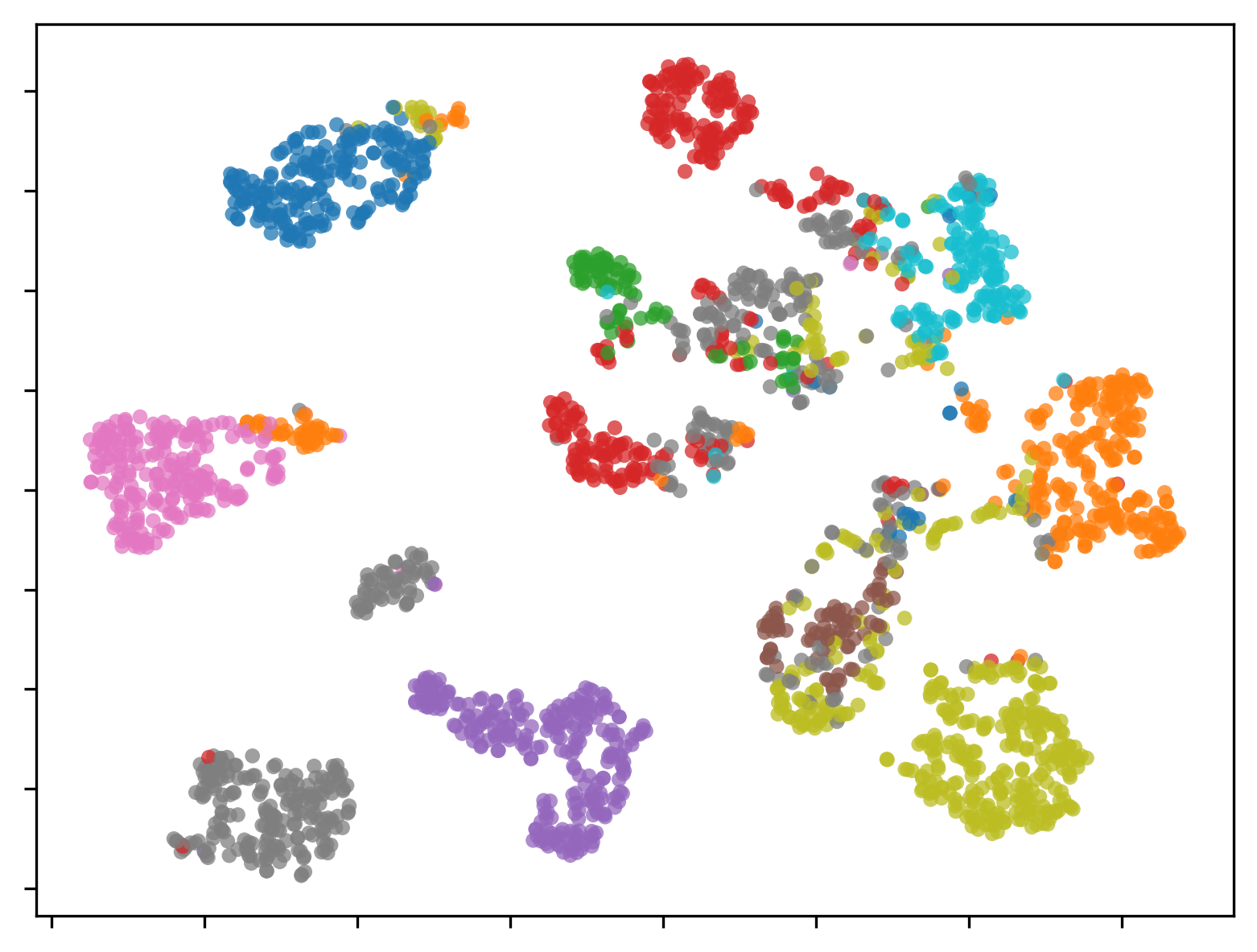}
        \put(-10, 20){\footnotesize \rotatebox{90}{PointMixer}}
        \put(75, 5){\footnotesize $e$=640}
    \end{overpic}
    \begin{overpic}[width=0.45\columnwidth, trim=2mm 0 2mm 0, clip]{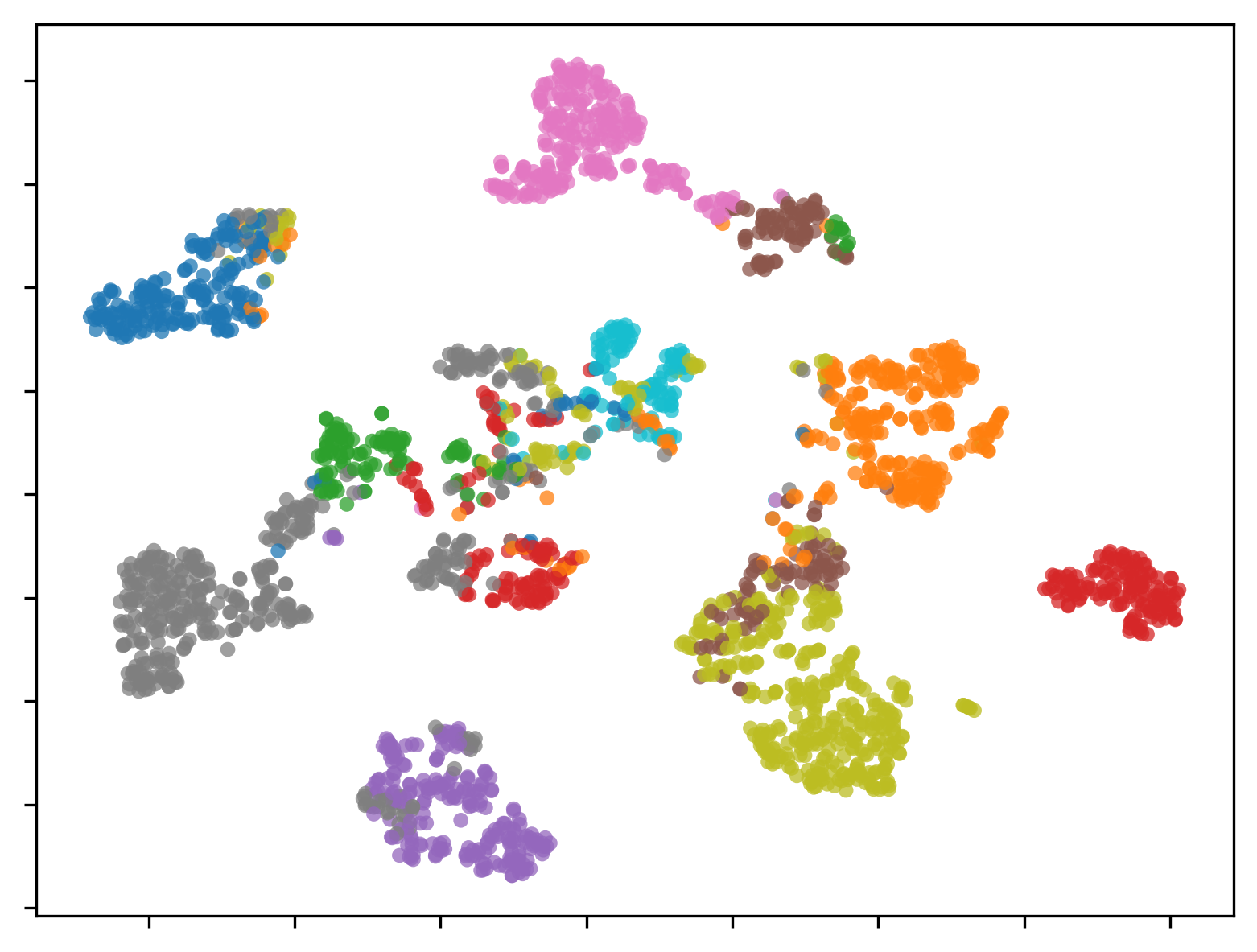}
        \put(75, 5){\footnotesize $e$=584}
    \end{overpic}
    \\
    \begin{overpic}[width=0.45\columnwidth, trim=2mm 0 2mm 0, clip]{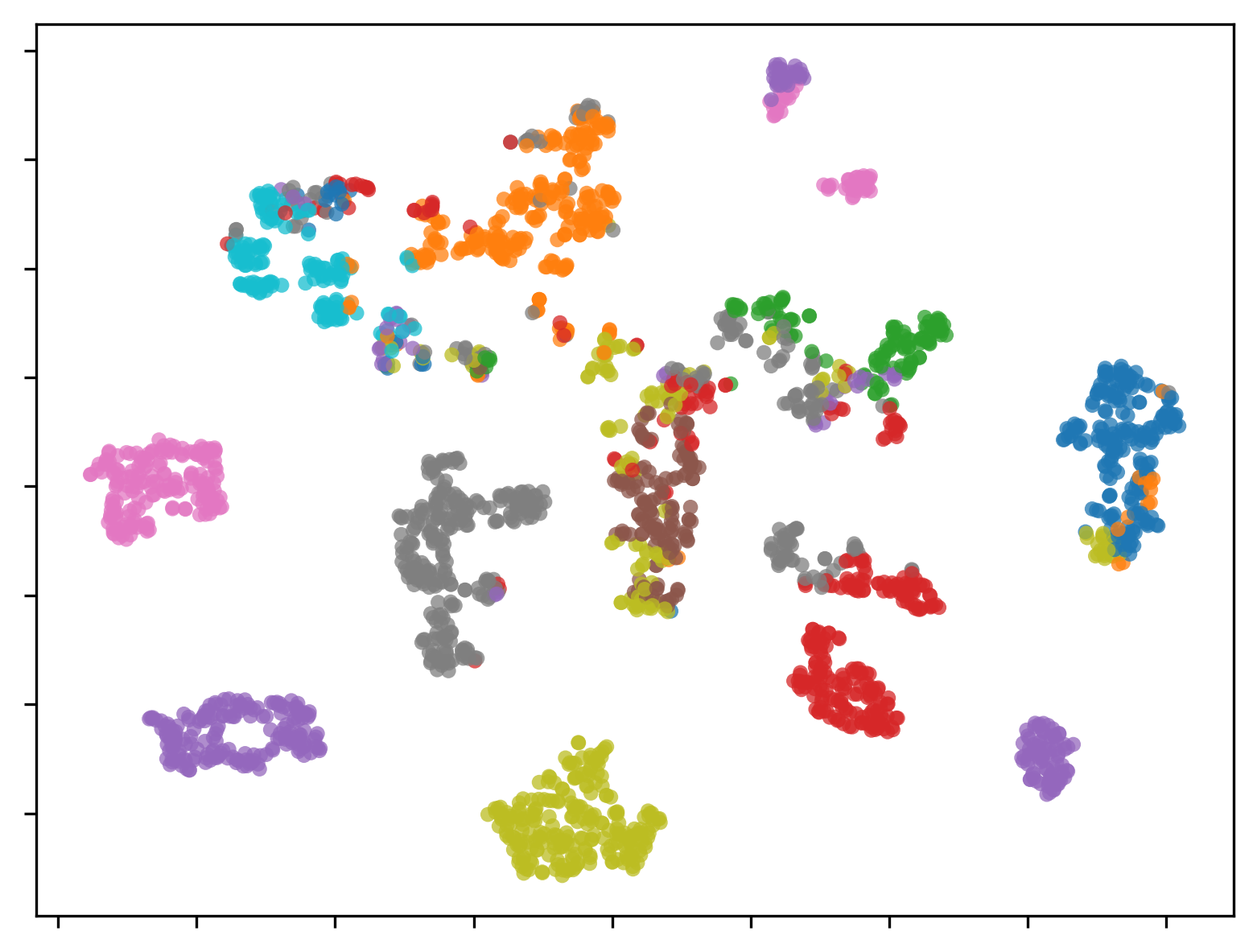}
        \put(-10, 7){\footnotesize \rotatebox{90}{PatchMixer (Ours)}}
        \put(75, 5){\footnotesize $e$=\textbf{617}}
    \end{overpic}
    \begin{overpic}[width=0.45\columnwidth, trim=2mm 0 2mm 0, clip]{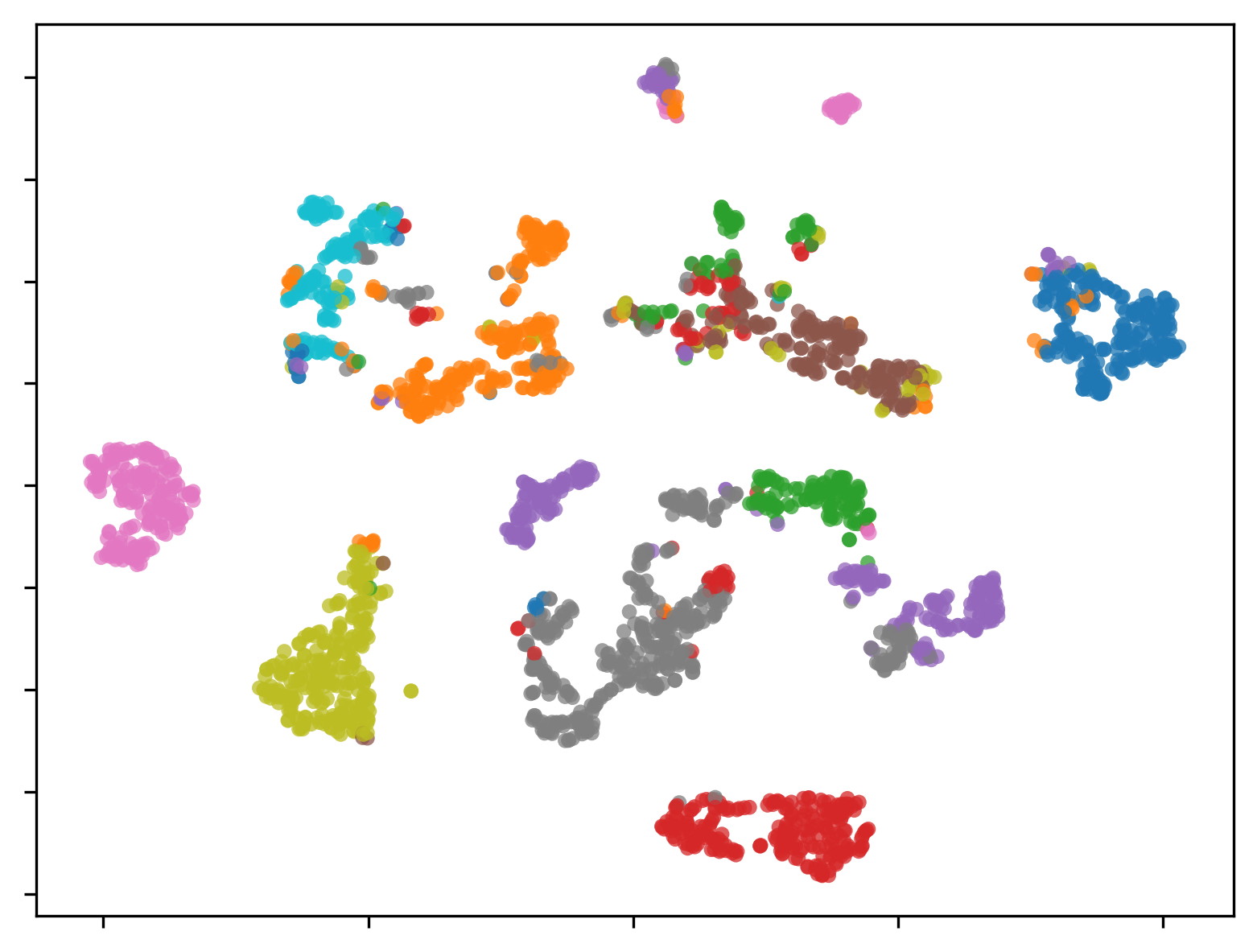}
        \put(75, 5){\footnotesize \contour{white}{$e$=\textbf{468}}}
    \end{overpic}
    \\
    \caption{
        Visualization of the t-SNE~\cite{tsne} embedding.
        Different object categories are displayed with different colors:
        {\color{box}{\Large$\bullet$}}~box,
        {\color{can}{\Large$\bullet$}}~can,
        {\color{banana}{\Large$\bullet$}}~banana,
        {\color{powerdrill}{\Large$\bullet$}}~powerdrill,
        {\color{scissors}{\Large$\bullet$}}~scissors,
        {\color{pear}{\Large$\bullet$}}~pear,
        {\color{dish}{\Large$\bullet$}}~dish,
        {\color{camel}{\Large$\bullet$}}~camel,
        {\color{mouse}{\Large$\bullet$}}~mouse,
        {\color{shampoo}{\Large$\bullet$}}~shampoo.
        $e$ denotes the number of wrongly classified shapes (the lower the better).
    }
    \label{fig:qual_clas_rr}
\end{figure}

\captionsetup{skip=0pt}
\begin{figure}[t!]
    \centering
    \vspace*{-4mm}

    \bigskip
    \begin{overpic}[width=1.0\columnwidth, trim=5mm 17mm 14mm 1mm, clip]{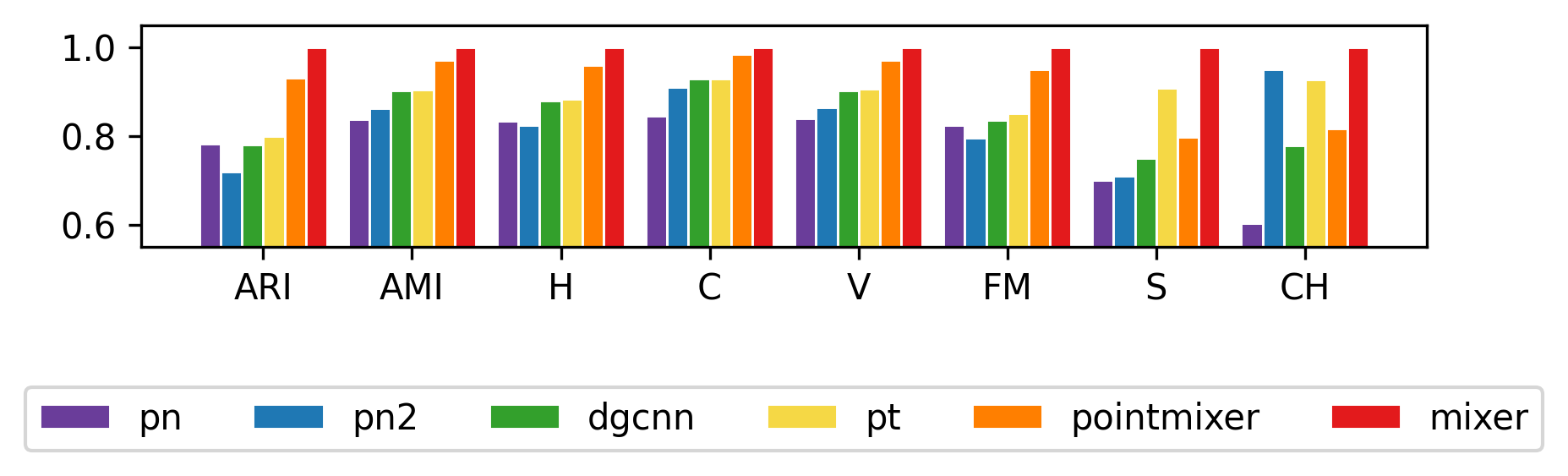}
        \put(36, 22){\footnotesize Synthetic-RS$\to$Real-RS}
    \end{overpic}
    \\
    \bigskip
    \begin{overpic}[width=1.0\columnwidth, trim=5mm 17mm 14mm 1mm, clip]{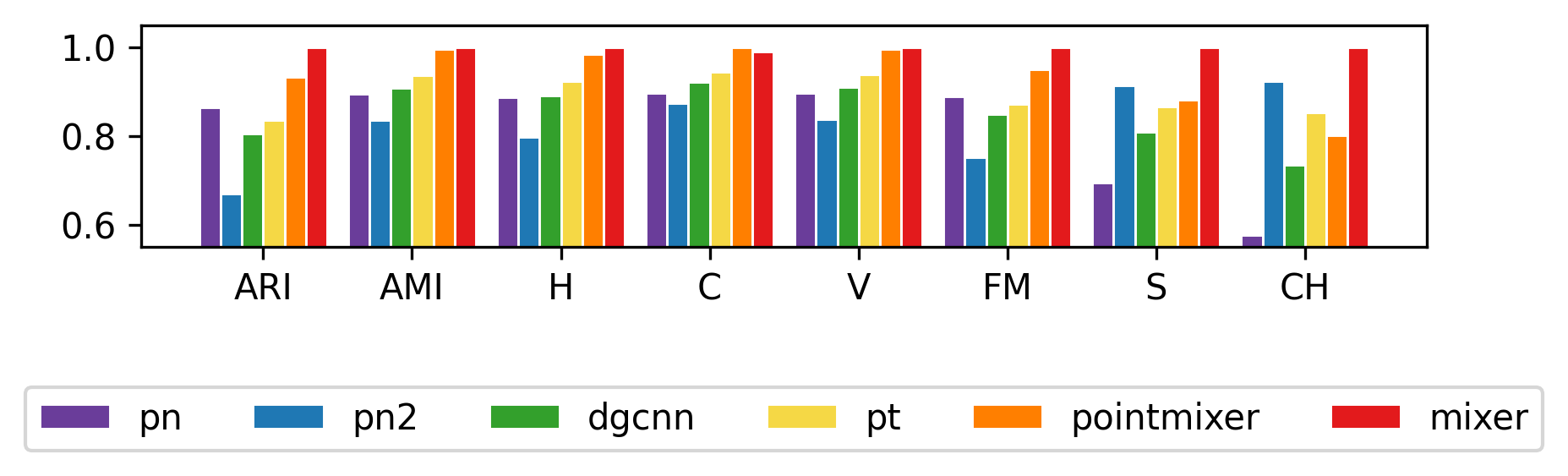}
        \put(37, 22){\footnotesize Synthetic-K$\to$Real-RS}
    \end{overpic}
    \\
    \caption{
        Comparison of clustering metrics (the higher the better).
        Methods key:
        {\color{pn}{\Large$\bullet$}}~PN~\cite{PN},
        {\color{pn2}{\Large$\bullet$}}~PN2~\cite{PN2},
        {\color{dgcnn}{\Large$\bullet$}}~DGCNN~\cite{DGCNN},
        {\color{pt}{\Large$\bullet$}}~PointTransformer~\cite{PointTransformer},
        {\color{pointmixer}{\Large$\bullet$}}~PointMixer~\cite{PointMixer},
        {\color{ours}{\Large$\bullet$}}~PatchMixer (Ours).
        Metrics key: 
        ARI: adjusted rand index,
        AMI: adjusted mutual information,
        H: homogeneity score, 
        C: completeness score,
        V: V-measure,
        FM: Fowlkes-Mallows score,
        S: silhouette coefficient,
        CH: Calinski-Harabasz index.
    }
    \label{fig:clustering_metrics}
\end{figure}

Table~\ref{tab:pointda} shows the classification results on PointDA~\cite{PointDAN} evaluated in terms of OA.
It contains three rows, one for each of the three subsets.
The top row shows the performance when the model is trained on the ModelNet (M) training split and tested on the ModelNet test split (grey background), and on the ShapeNet (S) and ScanNet (S$^\prime$) test splits (white background).
The other two rows show the results when the models is trained on the ShapeNet and ScanNet training splits, respectively.

PatchMixer outperforms the other approaches on two out of three splits and reaches the second-best performance on the third split.
When training on ModelNet, PatchMixer outperforms the other backbones and shows a 1.3\% improvement over PointMLP, which is the second-best method in this setting.
When training on ShapeNet, PatchMixer is the best method and performs on par with PointMLP.
When training on ScanNet, PN achieves the best performance, with a negligible gap compared to PointMLP and PatchMixer.
However, we suspect that the surprisingly good performance of PN in this particular setting can be attributed to an anomaly.
This anomaly arises from the fact that PN tends to overfit S$^\prime$ less compared to the other models because of (i) the significant difference in the number of parameters, with PN having only 800K parameters compared to an average of 5.44M parameters in other models, and (ii) the discrepancy in the amount of training data between S$^\prime$ and S (where PN performs poorly), with S$^\prime$ having three times less training samples than S (6K vs 17K samples).

\subsubsection{Qualitative results}
Fig.~\ref{fig:qual_clas_rr} shows 2D embeddings obtained with t-SNE~\cite{tsne} of global features learned for the shape classification task.
Training is performed on Synthetic-RS for the left column, and on Synthetic-K for the right one.
Inference is performed on Real-RS.
PatchMixer embeddings show clusters that are more compact and separated than other models, especially for the Banana, Pear, and Shampoo categories.
Scissors embeddings instead are splitted in two clusters, while all other approaches are able to map them in a single one.
Differences between models become more subtle as their performance get closer.

Fig.~\ref{fig:clustering_metrics} compares the same global features but using traditional clustering metrics, to allow for a better assessment of the differences between PatchMixer and competitors.
For an exhaustive analysis of intra- and inter-cluster statistics we considered different metrics.
Extrinsic clustering measures explicitly compare classification predictions with ground-truth annotations.
We considered the following six.
Adjusted Rand Index (ARI), which measures the similarity between the cluster assignments by making pair-wise comparisons.
Adjusted Mutual Information (AMI), which measures the agreement between the cluster assignments.
Homogeneity (H), which measures how many instances of a single class belong to a given cluster (similarly to Precision).
Completeness (C), which measures the instances of a given class that are assigned to the same cluster (similarly to Recall).
V-measure (V), which quantifies the correctness of the cluster assignments using conditional entropy analysis.
Fowlkes-Mallows score (FM), which measures the correctness of the cluster assignments using a geometric mean of pairwise Precision and Recall.
Intrinsic clustering measures do not require ground-truth annotations. 
We considered the following two.
Silhouette coefficient (S), which measures the ratio between inter- and intra-cluster distances.
Calinski-Harabasz index (CH), which measures the ratio between inter- and intra-cluster dispersion.
For all the metrics above, an higher score corresponds to a better performance.
The values reported in the histograms of Fig.~\ref{fig:clustering_metrics} are normalized so that the maximum value for each score is 1.
PatchMixer is the best method according to all the metrics in both scenarios, with the only exception of C in the SK$\to$RR split, where PointMixer performs slightly better.
PatchMixer's improvement margin in S and CH is quite significant.

\subsection{Part segmentation}\label{sec:part_segmentation}
Part segmentation can be formulated as a per-point classification problem.
Given a predefined set of parts, the goal is to predict the shape part to which it belongs to for each vertex of the observed point cloud.

\begin{table}[t!]
    \centering
    \caption[]{
    Part segmentation performance on PointSegDA~\cite{DefRec} evaluated in terms of mIoU.
    Dataset keys: A:~Adobe, F:~FAUST, M:~MIT, and S:~SCAPE.
    The structure follows Table~\ref{tab:graspnetpc}.
    }
    \label{tab:pointsegda}
    \smallskip
    \resizebox{1.0\columnwidth}{!}{%
        \begin{tabular}{lgcccc}
            \toprule
            Method & A$\to$A & A$\to$F & A$\to$M & A$\to$S & AvgTL \\
            \toprule
            PointNet \cite{PN} & 0.888 & 0.414 & 0.382 & 0.387 & 0.395 \\
            PointNet++ SSG \cite{PN2} & 0.905 & 0.406 & 0.430 & 0.374 & 0.403 \\
            DGCNN \cite{DGCNN} & 0.894 & 0.477 & 0.447 & 0.378 & 0.434 \\
            PointTransformer \cite{PointTransformer} & 0.912 & 0.516 & 0.513 & 0.465 & 0.498 \\
            PointMLP \cite{PointMLP} & 0.898 & 0.545 & 0.522 & 0.476 & \textbf{0.514} \\
            Ours & 0.854 & 0.552 & 0.515 & 0.476 & \textbf{0.514} \\
            \toprule
            Method & F$\to$F & F$\to$A & F$\to$M & F$\to$S & AvgTL \\
            \toprule
            PointNet \cite{PN} & 0.869 & 0.732 & 0.558 & 0.539 & 0.610 \\
            PointNet++ SSG \cite{PN2} & 0.885 & 0.683 & 0.628 & 0.647 & 0.653 \\
            DGCNN \cite{DGCNN} & 0.888 & 0.760 & 0.607 & 0.633 & 0.666 \\
            PointTransformer \cite{PointTransformer} & 0.890 & 0.746 & 0.625 & 0.626 & 0.666 \\
            PointMLP \cite{PointMLP} & 0.894 & 0.787 & 0.636 & 0.669 & \textbf{0.697} \\
            Ours & 0.874 & 0.712 & 0.615 & 0.689 & \underline{0.672} \\
            \toprule
            Method & M$\to$M & M$\to$A & M$\to$F & M$\to$S & AvgTL \\
            \toprule
            PointNet \cite{PN} & 0.744 & 0.737 & 0.595 & 0.493 & 0.608 \\
            PointNet++ SSG \cite{PN2} & 0.861 & 0.723 & 0.651 & 0.560 & 0.645 \\
            DGCNN \cite{DGCNN} & 0.789 & 0.721 & 0.646 & 0.593 & 0.653 \\
            PointTransformer \cite{PointTransformer} & 0.844 & 0.715 & 0.723 & 0.587 & 0.675 \\
            PointMLP \cite{PointMLP} & 0.838 & 0.729 & 0.702 & 0.728 & \textbf{0.720} \\
            Ours & 0.784 & 0.712 & 0.709 & 0.668 & \underline{0.696} \\
            \toprule
            Method & S$\to$S & S$\to$A & S$\to$F & S$\to$M & AvgTL \\
            \toprule
            PointNet \cite{PN} & 0.692 & 0.695 & 0.671 & 0.566 & 0.644 \\
            PointNet++ SSG \cite{PN2} & 0.825 & 0.668 & 0.771 & 0.659 & 0.699 \\
            DGCNN \cite{DGCNN} & 0.824 & 0.709 & 0.774 & 0.629 & 0.704 \\
            PointTransformer \cite{PointTransformer} & 0.799 & 0.782 & 0.646 & 0.799 & \textbf{0.743} \\
            PointMLP \cite{PointMLP} & 0.844 & 0.608 & 0.784 & 0.633 & 0.675 \\
            Ours & 0.798 & 0.743 & 0.799 & 0.654 & \underline{0.732} \\
            \bottomrule
        \end{tabular}
    }
\end{table}

\subsubsection{Dataset}
We use PointSegDA~\cite{DefRec} that contains human shapes collected from four datasets: Adobe Fuse 3D Characters (Adobe), FAUST~\cite{FAUST}, MIT~\cite{MIT}, and SCAPE~\cite{SCAPE}.
Different subsets contain the same person in different poses (MIT, SCAPE), different people in the same pose (Adobe), or different people in different poses (FAUST).
Each shape is provided with a manual annotation that segments it into the same eight parts: head, torso, upper-arm, forearm, hand, thigh, leg and foot.
Human shapes are provided in the form of point clouds either sampled from synthetic 3D meshes (Adobe, FAUST) or reconstructed from partial views (MIT, SCAPE).
Unlike GraspNetPC and PointDA, which only contain rigid objects related by rigid transformations, PointSegDA is more challenging since it contains non-rigid shapes affected by non-rigid deformations.

\begin{figure}[h!t]
    \centering

    \vspace{-2mm}
    \begin{overpic}[width=\columnwidth, trim=0 0 0 0, clip]{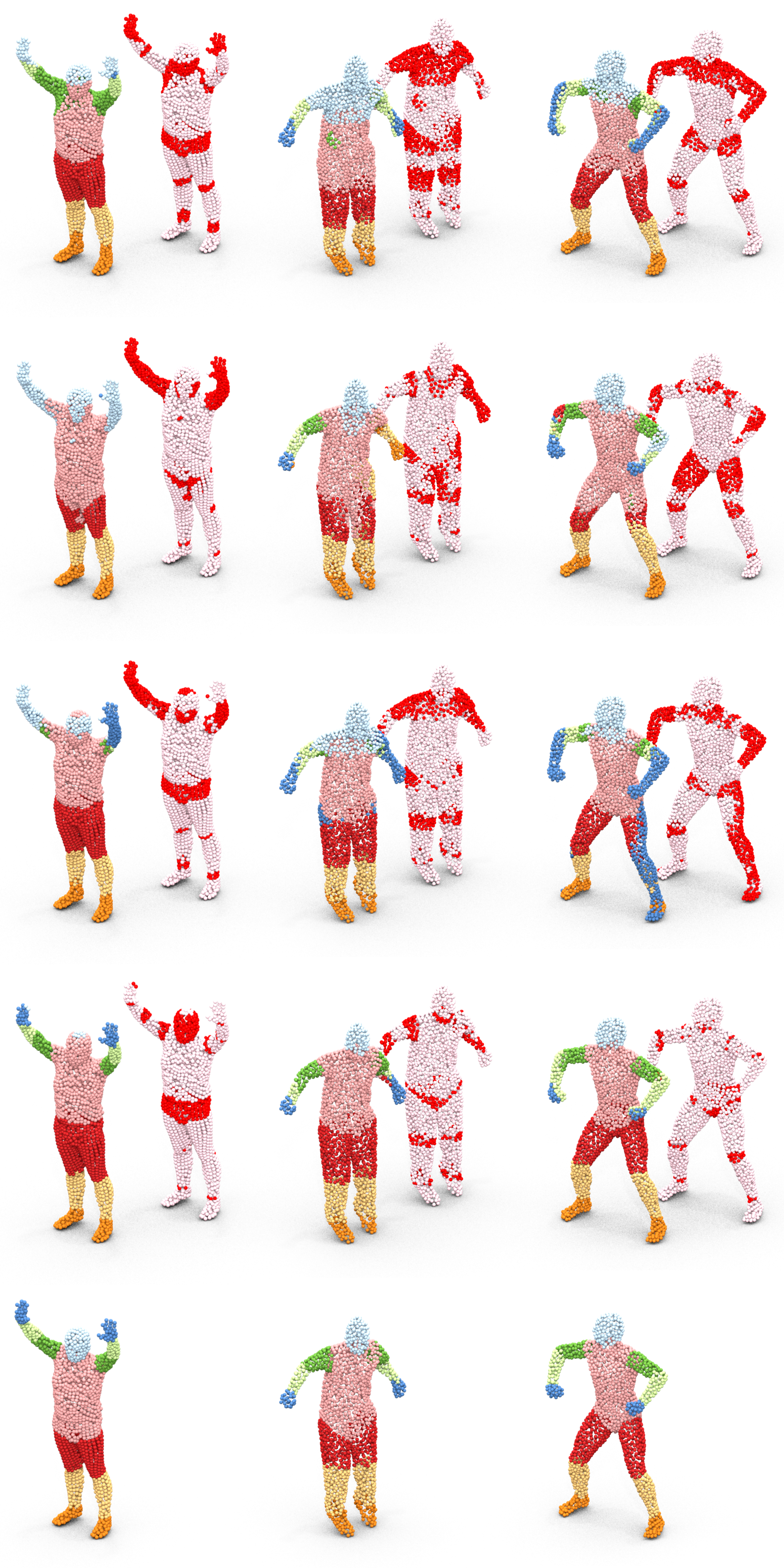}
        \put(0, 85){\rotatebox{90}{\footnotesize PointNet \cite{PN}}}
        \put(0, 62){\rotatebox{90}{\footnotesize PointNet++ \cite{PN2}}}
        \put(0, 44){\rotatebox{90}{\footnotesize DGCNN \cite{DGCNN}}}
        \put(0, 27){\rotatebox{90}{\footnotesize Ours}}
        \put(0, 9){\rotatebox{90}{\footnotesize GT}}
        \put(7, 81){\footnotesize $e$=641}
        \put(23, 81){\footnotesize $e$=886}
        \put(40, 81){\footnotesize $e$=641}
        \put(7, 60){\footnotesize $e$=581}
        \put(23, 60){\footnotesize $e$=565}
        \put(40, 60){\footnotesize $e$=537}
        \put(7, 39){\footnotesize $e$=605}
        \put(23, 39){\footnotesize $e$=549}
        \put(40, 39){\footnotesize $e$=640}
        \put(7, 19){\footnotesize $e$=\textbf{453}}
        \put(23, 19){\footnotesize $e$=\textbf{367}}
        \put(40, 19){\footnotesize $e$=\textbf{218}}
        \put(2, 0){\footnotesize Adobe$\to$FAUST}
        \put(21, 0){\footnotesize Adobe$\to$MIT}
        \put(36, 0){\footnotesize Adobe$\to$SCAPE} 
    \end{overpic}

    \vspace{1mm}
    \caption{
        Visualization of few part segmentation results and relative error maps.
        Part segmentation color key:
        {\color{head}{\Large$\bullet$}}~head,
        {\color{torso}{\Large$\bullet$}}~torso,
        {\color{upperarms}{\Large$\bullet$}}~upper-arms,
        {\color{forearms}{\Large$\bullet$}}~forearms,
        {\color{hands}{\Large$\bullet$}}~hands,
        {\color{thighs}{\Large$\bullet$}}~thighs,
        {\color{legs}{\Large$\bullet$}}~legs,
        {\color{foot}{\Large$\bullet$}}~foot.
        Error map color key:
        {\color{wrong}{\Large$\bullet$}}~wrong,
        {\color{correct}{\Large$\bullet$}}~correct.
        $e$ denotes the number of points wrongly segmented (the lower the better).
        Rows 1--4 represent different network designs.
        The last row shows the ground truth part segmentation.
    }
    \vspace{-2mm}
    \label{fig:qual_segm}
\end{figure}
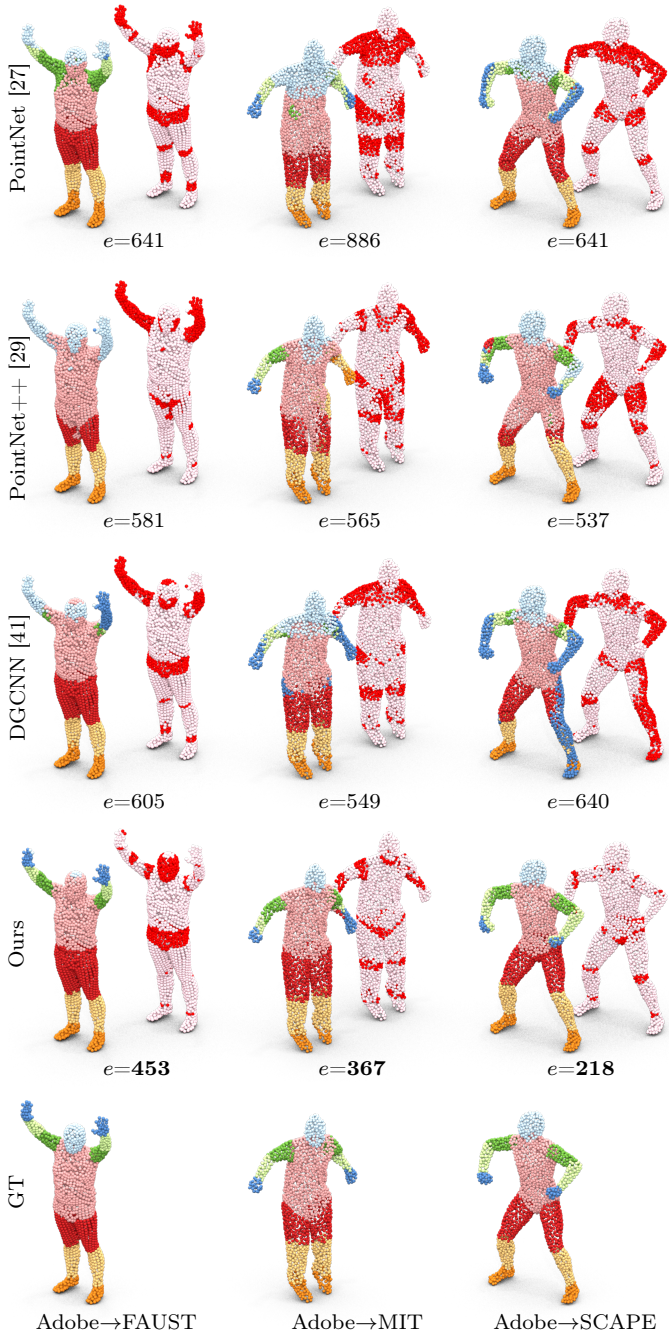

\subsubsection{Network configuration}
Input shapes are sampled to $V=2048$ vertices.
We use $P=256$ patches.
Because part segmentation requires point-level predictions, to extract more local patches we used kNN instead of ball query to extract $S=64$ samples for each patch.
We use patch masking to drop around 30\% of patches.
Each patch is sampled to $S=128$ samples.
The patch embedding is implemented as an MLP producing features of size $F=1024$.
Because the data available in PointSegDA~\cite{DefRec} are orders of magnitude less that the ones provided by the classification benchmarks, we reduce the capacity of our model by decreasing the hidden feature dimension for the token and channel mixers to $F_\text{T}=F_\text{C}=64$.
The depth of the model is set to $D=4$.

\subsubsection{Training parameters}
We use the same training parameters as those used for classification.
In addition to the data augmentation described above, we translate each shape around its barycenter by a random factor in the range ($-$0.1, 0.1).

\subsubsection{Quantitative results}
Table~\ref{tab:pointsegda} shows the part segmentation performance on PartSegDA~\cite{DefRec} evaluated in terms of the mean Intersection-over-Union (mIoU) metric.
It contains four rows, in each row we train on the training split of a dataset and test on the test split of the same dataset (same domain, grey background) and on the test split of the other three datasets (TL, white background).
By considering the splits separately, our approach either reach the best performance or the second best in each of the four cases. 
Overall, by averaging the AvgTL mIoU over the four splits, we obtain 65.35 for PatchMixer and 65.15 for PointMLP.
We achieve a slightly better performance than PointMLP, the current state-of-the-art deep architecture for 3D point cloud processing.
This result is particularly remarkable since our model uses only around 1/4 of the parameters used by PointMLP: 4.2M compared to 16.7M parameters.

\subsubsection{Qualitative results}
Fig.~\ref{fig:qual_segm} shows some qualitative examples.
The first four rows report the results obtained with PointNet~\cite{PN}, PointNet++~\cite{PN2}, DGCNN~\cite{DGCNN}, and PatchMixer. 
The last row shows the ground-truth part segmentation.
Different colors represent different human shape parts.
Behind each segmentation result we show its error map, where wrongly segmented vertices are colored in red.
Our approach provides the more accurate results and it often does not miss any part, with the exception of ``hand''.
The other approaches lack accuracy and tend to merge together semantically different parts.

\subsection{Ablation study}
We perform an ablation study using the shape classification task on a subset of GraspNetPC~\cite{GAI} in Table~\ref{tab:ablation_grasp}.

\begin{table}[t!]
    \centering
    \tabcolsep 3pt
    \caption[]{
        Ablation study on a subset of GraspNetPC~\cite{GAI}.
        Keys: $R$: patch radius, TM: Token Mixer, Att.: Attention, $D$: depth of the mixer modules, $B$: batch size.
    }
    \label{tab:ablation_grasp}
    \resizebox{1.0\columnwidth}{!}{%
        \begin{tabular}{cccccgcgcc}
            \toprule
            $R$ & TM & Att. & $D$ & $B$ & RK$\to$RK & RK$\to$RR & RR$\to$RR & RR$\to$RK & AvgTL \\
            \toprule
            0.1 & \cmark & \cmark & 1 & 32 & 0.981 & 0.682 & 0.941 & 0.841 & 0.762 \\
            0.3 & \cmark & \cmark & 1 & 32 & 0.994 & 0.834 & 0.945 & 0.897 & \textbf{0.866} \\
            0.5 & \cmark & \cmark & 1 & 32 & 0.987 & 0.826 & 0.929 & 0.866 & 0.846 \\
            1.0 & \cmark & \cmark & 1 & 32 & 0.990 & 0.813 & 0.940 & 0.888 & 0.851 \\
            \midrule
            0.3 &  &  & 1 & 32 & 0.974 & 0.807 & 0.925 & 0.868 & 0.838 \\
            0.3 & \cmark &  & 1 & 32 & 0.993 & 0.828 & 0.940 & 0.892 & 0.860 \\
            0.3 & \cmark & \cmark & 1 & 32 & 0.994 & 0.834 & 0.945 & 0.897 & \textbf{0.866} \\
            \midrule
            0.3 & \cmark & \cmark &  1 & 16 & 0.985 & 0.839 & 0.952 & 0.909 & 0.874 \\
            0.3 & \cmark & \cmark & 16 & 16 & 0.994 & 0.863 & 0.951 & 0.945 & \textbf{0.904} \\
            \midrule
            0.3 & \cmark & \cmark & 1 &  4 & 0.095 & 0.105 & 0.143 & 0.117 & 0.111 \\
            0.3 & \cmark & \cmark & 1 &  8 & 0.285 & 0.209 & 0.096 & 0.101 & 0.155 \\
            0.3 & \cmark & \cmark & 1 & 16 & 0.978 & 0.837 & 0.941 & 0.947 & \textbf{0.892} \\
            0.3 & \cmark & \cmark & 1 & 32 & 0.994 & 0.834 & 0.945 & 0.897 & 0.866 \\
            \bottomrule
        \end{tabular}
    }
    \vspace{-2mm}
\end{table}

\subsubsection{Local vs global receptive field}
Patch size depends on the radius $R$.
Our experiments show that small patches ($R=0.3$) led to better transfer learning performance than large ones (e.g.~$R=0.5, 1.0$).
Reducing the radius further ($R=0.1$) has a detrimental effect because it cannot capture enough geometric context around the patch centroid.
Decomposing the input shape in a collection of small patches allows the network to learn features more informative of local geometric details, while large patches lead to more global features which are more sensitive to intra-class variations and occlusions.
In addition, the patch masking augmentation works better when removing small patches rather than big ones, since by randomly removing small patches we can better emulate the acquisition artifacts real-world data are affected by without compromising the global shape of the object.

\subsubsection{Does mixing patches help?}
One of the main novelties of the proposed architecture is the introduction of the attentive token mixer module.
To assess its contribution, we compare PatchMixer against two baselines where:
i) the attentive token mixer is removed, and
ii) the attention module inside the token mixer is removed, i.e.~Eq.~\eqref{eq:attentive_token_mixer} is replaced by Eq.~\eqref{eq:token_mixer}.
The baseline using only the channel mixer module reach an OA of 83.8 in AvgTL.
Mixing tokens following Eq.~\eqref{eq:token_mixer}, without using the attention module, improves the AvgTL by 2.2\%.
Introducing the attention module with Eq.~\eqref{eq:attentive_token_mixer} further boost the performance by 0.6\%.
Mixing tokens incorporates contextual information learning more discriminative features, while the attention module learns to consider only significant patches, ignoring outliers.

\subsubsection{Depth of the mixing layers}
Increasing the depth of the mixing layers increases the capacity of the model and has a beneficial effect when the cardinality of the training dataset is big enough.
In this case, it improves the OA by 3.0\% in AvgTL.
Despite the ablation results suggest that increasing the depth of the model lead to an increase in performance, the experiments presented throughout the paper are performed with $D=1$ to keep the number of PatchMixer parameters low.

\subsubsection{Mini-batch size}
When using very small batch size of 4 or 8 shapes, PatchMixer performs very poorly.
Increasing the batch size the performance improves.
We obtained the best performance for $B=16$.
A typical choice of the batch size for the other architecture designs we are comparing with is 32.
To make the comparison fair we choose $B=32$ throughout all the experiments presented in this paper.

\section{Conclusions}\label{sec:conclusions}

We presented a novel network design for 3D point cloud processing that is intrinsically effective in generalization across data distributions.
Compared to existing architectures, PatchMixer can better generalise across domains, while also being as effective as state-of-the-art methods in the same domain scenario.
We also provided a comprehensive transfer learning evaluation that was missing in the literature, which we hope it will be adopted as common practice to better assess the effectiveness of different methods.

\section*{Acknowledgements}
This work was supported by the European Union’s Horizon Europe research and innovation programme under grant agreement No 101058589 (AI-PRISM), and by the PNRR project FAIR - Future AI Research (PE00000013), under the NRRP MUR program funded by the NextGenerationEU.

\bibliographystyle{elsarticle-harv}
\bibliography{biblio}

\end{document}